\documentclass[journal]{IEEEtran}
\usepackage{enumitem}
\usepackage{algorithmic}
\usepackage{algorithm}
\usepackage{siunitx}
\usepackage{amssymb}
\usepackage{booktabs}
\usepackage{xcolor}
\usepackage{mathtools}
\usepackage{cuted}

\usepackage[1]{editing}

\usepackage[subrefformat=parens,labelformat=parens]{subfig}

\ifCLASSINFOpdf
   \usepackage[pdftex]{graphicx}
\else
\fi
\usepackage{amsmath}
\usepackage{array}

\usepackage{tabularx,booktabs}

\begin{document}

\title{Efficient Covariance Matrix Reconstruction with Iterative Spatial Spectrum Sampling}
%
%
%

\author{Saeed~Mohammadzadeh,~\IEEEmembership{Member,~IEEE,}
        V\'{\i}tor~H.~Nascimento,~\IEEEmembership{Senior,~IEEE,}
        Rodrigo~C.~de~Lamare,~\IEEEmembership{Senior,~IEEE,}
        and~Osman~Kukrer,~\IEEEmembership{Senior,~IEEE} \vspace{-0.75em}}

\maketitle

\begin{abstract}
This work presents a cost-effective technique for designing robust adaptive beamforming algorithms based on efficient covariance matrix reconstruction with iterative spatial power spectrum  (CMR-ISPS). The proposed CMR-ISPS approach reconstructs the interference-plus-noise covariance (INC) matrix based on a simplified maximum entropy power spectral density function that can be used to shape the directional response of the beamformer. Firstly, we estimate the directions of arrival (DoAs) of the interfering sources with the available snapshots. We then develop an algorithm to reconstruct the INC matrix using a weighted sum of outer products of steering vectors whose coefficients can be estimated in the vicinity of the DoAs of the interferences which lie in a small angular sector. We also devise a cost-effective adaptive algorithm based on conjugate gradient techniques to update the beamforming weights and a method to obtain estimates of the signal of interest (SOI) steering vector from the spatial power spectrum. The proposed CMR-ISPS beamformer can suppress interferers close to the direction of the SOI by producing notches in the directional response of the array with sufficient depths. Simulation results are provided to confirm the validity of the proposed method and make a comparison to existing approaches. 
\end{abstract}


%
\IEEEpeerreviewmaketitle

\section{Introduction}
%
%
%
%
\IEEEPARstart{A}{daptive} beamforming is a fundamental signal processing technique that has been widely used in wireless communications, radar, sonar, and other fields \cite{van2004detection} to improve the quality of signal reception. The standard minimum variance distortionless response (MVDR) \cite{capon1969high} beamformer is a well-known adaptive beamformer that assumes accurate knowledge of the antenna array, the actual array manifold, and that there is no desired signal component in the training samples. MVDR is used to suppress interference and noise with a distortionless desired signal. However, imprecise knowledge of steering vectors (SV) and training data can lead to performance degradation. Moreover, MVDR beamforming cannot deal with signal mismatches because of its sensitivity to errors due to direction-of-arrival (DoA) estimation, array calibration, wavefront distortion, and incoherent local scattering. 

Many robust adaptive beamforming techniques that can mitigate the effects of model mismatches and improve the robustness of beamformers have been reported. In general, robust adaptive beamforming algorithms can be classified as follows: diagonal loading (DL) techniques, eigenspace-based techniques, uncertainty-set-based techniques, and interference-plus-noise covariance (INC) matrix reconstruction-based techniques. Diagonal loading techniques \cite{mestre2006finite,kukrer2014generalised,rccm,rrelay,rmmse,rstap} increase robustness against the sensitivity of the desired signal mismatches and finite training sample effects by introducing a loading factor that is added to the diagonal entries of the sample covariance matrix. However, the main drawback of this approach is that it is difficult to choose the optimal DL factor in different scenarios. The eigenspace-based technique is another type of robust adaptive beamforming approach, which is obtained by projecting the nominal steering vectors onto the signal-plus-interference subspace to eliminate the arbitrary steering vector mismatches of the signal-of-interest (SOI) \cite{huang2012modified,jia2013robust,rjio}. However, the performance of the eigenspace-based beamformer will degrade drastically under low signal-to-noise ratios (SNR) \cite{xie2014fast}. Uncertainty set algorithms such as the worst-case performance optimization algorithm \cite{vorobyov2003robust,yu2010robust,nai2011iterative} and the linear programming algorithm \cite{jiang2014robust} obtain an optimal solution by establishing an ellipsoidal or spherical uncertainty constraint on the SOI steering vector. However, their performance is affected by the uncertainty parameter set and it is difficult to select the optimal factor in practice \cite{yang2017modified}. Moreover, these algorithms do not eliminate the SOI component from the sample covariance matrix, which causes performance loss at high SNR.

The robust adaptive beamforming methods described above are intended to estimate the desired signal steering vector or the sample covariance matrix. Although they can improve the robustness of beamformers, they all suffer from severe performance degradation at high input SNR. To address the performance degradation caused by high input SNR, robust adaptive beamforming methods based on the INC matrix reconstruction have been developed in the last decade \cite{mallipeddi2011robust,gu2012robust,yuan2017robust,mohammadzadeh2018modified,ruan2014robust,huang2015robust,oskpme,lrcc,chen2018robust,mohammadzadeh2019robustcssp,zheng2018covariance,mohammadzadeh2018adaptive,zhang2021adaptive,sun2021robust,mohammadzadeh2022robust,Li2022Jul,bbprec,Mohammadzadeh2022Dec,Luo2022,9913641,9991136,mbthp,rmbthp,rsrbd,rsthp}. The study in \cite{mallipeddi2011robust} first employed the standard Capon beamformer to estimate the interference steering vectors and then reconstruct the INC matrix, but the power of the interference and desired signal steering vectors is not precisely estimated. To improve the estimation performance of the sample covariance matrix, the Capon power spectrum has been used to reconstruct the INC matrix in \cite{gu2012robust} by integrating over an angle sector excluding that of the SOI, while the SOI  is estimated by solving a quadratically constrained quadratic programming (QCQP) problem with high computational complexity. Although this method shows good performance, it is sensitive to large DoA mismatches and arbitrary amplitude and phase perturbation errors \cite{yuan2017robust,mohammadzadeh2018modified}. Low-complexity shrinkage-based mismatch estimation (LOCSME) and the sparsity of the source distribution are used in \cite{ruan2014robust} to significantly reduce the complexity. However, the performance of LOCSME is degraded in high interference power. In order to reconstruct the INC matrix, an annular uncertainty set was used in \cite{huang2015robust} to constrain the interferers. This method performs very similarly to the beamformer in \cite{gu2012robust}. However, because the INC matrix is reconstructed by integrating over a complex annular uncertainty set, it has a high computational complexity. Using the covariance matrix taper technique, a partial power spectrum sampling method has been studied in \cite{zhang2016interference} to reconstruct the INC matrix with low computational complexity, however, the required number of array elements is relatively large. An approach inspired by the weighted subspace-fitting-based INC matrix reconstruction beamformers was proposed in \cite{chen2018robust}, which is especially designed to mitigate the effect of sensor position errors. Furthermore, the work in \cite{mohammadzadeh2019robustcssp} proposed a low-complexity beamformer in which the square of the sample covariance matrix is used in the Capon estimator and is estimated based on a correlation sequence.
In \cite{zheng2018covariance}, the INC matrix reconstruction and the desired signal  estimation are based on a similar procedure to that of \cite{gu2012robust} and \cite{li2003robust}. However, ad-hoc parameters influence the calculation accuracy of the interference.
The approach in \cite{mohammadzadeh2018adaptive} employs the beamformer output power to jointly estimate the theoretical INC matrix and the mismatched  using the eigenvalue decomposition of the received signal covariance matrix.
The beamformer in \cite{zhang2021adaptive} employs a method devised by separating the SOI component from the training data with a blocking matrix, while the SOI steering vector is estimated as the principal eigenvector of the SOI covariance matrix. Then, the quasi-INC matrix is calculated using the SOI-free data.

The results of \cite{gu2012robust} demonstrate that the resulting Capon beamformer allows for good performance in the case of SOI array errors. However, the analysis did not account for typical interference array errors or arbitrary SOI array  mismatches \cite{somasundaram2014degradation}. In addition, the accuracy of the Capon spatial spectrum degrades severely when coherent signals (with line spectra) exist \cite{wang2016robust}. To avoid this problem, a recent efficient robust adaptive beamforming technique has been introduced in \cite{mohammadzadeh2021robust} based on the autocorrelation sequence of a random process in which the INC matrix is reconstructed directly and without the need to estimate the power of the interferers and their arrays. In \cite{sun2021robust}, an algorithm that employs the gradient vector and the INC matrix reconstruction by estimating the interference steering vectors and their powers is proposed. \\
\indent Although the above-mentioned INC matrix reconstruction approaches considerably enhance beamforming performance, they require numerical integration using a large number of sampling points, increasing beamforming complexity. Moreover, they assume the number of sources and corresponding DoAs as prior information. In this work, unlike previous methods, the DoAs of the interfering sources are estimated over the available snapshots while we assume that the angular sector corresponding to the desired signal is available. Then, the beamformer is designed to place a null region that spans the directions in which the interfering sources could be located. In addition, a cost-effective  approach to INC matrix reconstruction is devised based on a simplified  power spectral density function that can be used to shape the directional response of the beamformer. Furthermore, an iterative technique to compute this matrix is developed based on conjugate gradients which facilitates the fast calculation of the beamformer weights given the DoAs of the interferers. Moreover, a method is developed to obtain estimates of the SOI steering vector from the spatial power spectrum. The proposed CMR-ISPS beamformer can efficiently suppress interference signals from interfering sources close to the SOI direction by producing notches in the directional response of the array with sufficient widths and depths. The contributions of this paper are summarized as:
   \begin{itemize}
   \item A procedure that enables the proposed CMR-ISPS algorithm to adapt itself to the motion of interferers by determining the DoA and corresponding angular sector region where the null in the beampattern is to be created.
   \item A cost-effective technique, denoted as CMR-ISPS, for  designing  robust adaptive beamforming  algorithms based on efficient INC matrix reconstruction with iterative spatial spectrum sampling. 
        \item An approach to update the beamforming weights of the proposed CMR-ISPS algorithm with a reduced cost as it does not explicitly form the covariance matrices, relying instead on low-cost iterative techniques.
   \end{itemize}

The remainder of this paper is structured as follows. Section \ref{sec2} describes the signal model and  background. Section \ref{sec3} presents spectral and interference DoA estimation techniques. In Section \ref{sec4} the INC matrix reconstruction of the proposed  CMR-ISPS algorithm is presented. Section \ref{sec5} presents the estimation of the desired signal. In Section \ref{sec6} the steps of the proposed CMR-ISPS algorithm are detailed. Section \ref{sec7} analyzes properties of the maximum entropy INC matrix,  of the array gain, and the computational complexity of CMR-ISPS. Simulation results are described in Section \ref{sec8}. Finally, conclusions are drawn in Section \ref{sec9}.

\section{Problem background and the signal model}\label{sec2}

Consider an array of $ M $ sensors with interelement spacing $d $ that receives narrowband signals from multiple sources located in the far field. The $ {M \times1} $ complex received signal vector at time instant $t$ can be modeled as
\begin{align}\label{x}
\mathbf{{x}}(k)=s_0(k)\mathbf{a}_0+\sum_{l=1}^L s_l(k) \mathbf{a}_l +\mathbf{{n}}(k),
\end{align}
where $s_l(k), l=0,1,\ldots, L$ and $\mathbf{a}_l$ represent the received narrowband signals and the manifold vectors (or s). The manifold vectors are functions of the location parameters of the sources (e.g., their directions of arrival) and of the array geometry. Here we use the subscripts $l=0$ and $l=1,\cdots,L$ to indicate the components related to the SOI and the interfering signals respectively. The vector $\mathbf{n}(k)$ denotes the additive white Gaussian noise with power $\sigma^2_\text{n}$. In this work, we assume that the SOI, interferences, and noise are statistically independent of each other. Consequently, the covariance matrix of the received signal vector may be expressed as \vspace{-0.4em}
\begin{align} \label{Theoretical R}
\mathbf{R}=\mathcal{E}\lbrace  \mathbf{x}(k)\mathbf{x}^\mathrm{H}(k) \rbrace &=\sigma^2_0 \mathbf{a}_0 \mathbf{a}_0^\mathrm{H}+\sum_{l=1}^L \sigma^2_l \mathbf{a}_l \mathbf{a}_l^\mathrm{H}+\sigma^2_\text{n} \mathbf{I},
\end{align} 
where the notations $\mathcal{E} \lbrace \cdot \rbrace$ and $(\cdot)^\mathrm{H}$ stand for the expectation operator and conjugate transpose respectively, $\sigma^2_0$ and $\lbrace \sigma^2_l \rbrace_{l=1}^L$
are the powers of the SOI and interference signals, and the matrix $\mathbf{I}$ denotes the identity matrix. The desired signal covariance matrix and the INC matrix are denoted by $\mathbf{R}_\mathrm{s}$ and $\mathbf{R}_\mathrm{in}=\mathbf{R}_\mathrm{i}+\mathbf{R}_\mathrm{n}=\sum_{l=1}^L \sigma^2_l \mathbf{a}_l \mathbf{a}_l^\mathrm{H}+\sigma^2_\text{n} \mathbf{I}$, respectively. It should be noted that $\mathbf{R}_\mathrm{i}$ and $\mathbf{R}_\mathrm{n}$ depict the interference and noise covariance matrices, respectively.

The adaptive beamformer aims to allow the SOI to pass through without any distortion while the interference and noise are suppressed as much as possible, thereby maximizing the output signal-to-interference-plus-noise ratio (SINR). The standard Capon beamformer can be formulated as
\begin{align}\label{MVDR}
\underset{{\mathbf{w}}}{\operatorname{min}}\ \mathbf{w}^\mathrm{H} \mathbf{R}_{\mathrm{in}}\mathbf{w}\ \hspace{.4cm} \mathrm{s.t.} \hspace{.4cm} \mathbf{w}^\mathrm{H} \mathbf{a}_0=1,
\end{align}
with the solution
\begin{align}\label{optimal wegight vector}
\mathbf{w}_{\mathrm{opt}}={\mathbf{R}_{\mathrm{in}}^{-1} \mathbf{a}_0} / {\mathbf{a}_0^\mathrm{H} \mathbf{R}_{\mathrm{in}}^{-1}\mathbf{a}_0},
\end{align}
where $ \mathbf{{w}}=[w_{1}, \ \cdots, \ w_{M} ]^{T} $. The output SINR is defined as
\begin{align}\label{SINR}
\mathrm{SINR} \triangleq {\sigma^{2}_0 |\mathbf{{w}}^\mathrm{H} \mathbf{{a}}_0|^2 } / {\mathbf{{w}}^\mathrm{H} \mathbf{{R}}_{\mathrm{in}}\mathbf{{w}}}.
\end{align}
The general format of the array steering vector corresponding to the DoA of the signals in this manuscript is defined as
$\mathbf{a}(\theta)=[1, e^{-j \theta}, \cdots, e^{-j  (M-1)\theta} ]^T$, where $\theta=(2 \pi d/ \lambda) \sin \phi$ (assuming half-wavelength sensor spacing, $d=\lambda/2 $ ), $\phi$ is the angle of arrival and $(\cdot)^T$ denotes transposition.
However, in practical applications, the actual  of the SOI, $ \mathbf{a}_0 $, and the actual INC matrix, $ \mathbf{R}_{\mathrm{in}}$ are unavailable. Therefore, $\mathbf{a}_0 $ is usually replaced by the assumed , $\bar{\mathbf{a}}_0$, and in some situations $\mathbf{R}_{\mathrm{in}}$ can be replaced by the sample covariance matrix \cite{van2004detection} \vspace{-.7em}
\begin{align}\label{SCM}
\hat{\mathbf{R}}=({1}/ {K})\sum_{k=1}^{K} \mathbf{{x}}(k)\mathbf{{x}}^\mathrm{H}(k),
\end{align}
where \add{$K$} is the number of snapshots. Even though $\hat{\mathbf{R}}$ is an approximation for $\mathbf{R}$, the beamformer obtained by using the sample covariance matrix in \eqref{SCM} is known as the sample matrix inversion (SMI) technique and given by
\begin{align}\label{SMI wegight vector}
\mathbf{w}_{\mathrm{SMI}}={\hat{\mathbf{R}}^{-1} \bar{\mathbf{a}}_0} / {\bar{\mathbf{a}}_0^\mathrm{H} \hat{\mathbf{R}}^{-1}\bar{\mathbf{a}}_0}.
\end{align}
{As it is well known, the SMI beamformer is sensitive to mismatches, especially at high SNRs \cite{ruan2014robust}. With the beamformer in \eqref{SMI wegight vector}, the output power of the SMI beamformer is given by}
\begin{align}\label{Power SMI}
\hat{P}=\mathbf{w}_{\mathrm{SMI}}^\mathrm{H} \hat{\mathbf{R}}\mathbf{w}_{\mathrm{SMI}}={1} / {\bar{\mathbf{a}}_0^\mathrm{H} \hat{\mathbf{R}}^{-1}\bar{\mathbf{a}}_0},
\end{align}
which is the well-known Capon power spectrum. The focus of some recent adaptive beamforming methods is on using the Capon spectral estimation in \eqref{Power SMI} to compute the power from the direction of the signal, which has some disadvantages \cite{mohammadzadeh2019robustcssp,wang2016robust,zhang2013robust}. Based on the Capon spatial spectrum estimator \cite{capon1969high}, $ 1/\mathbf{a}_0^\mathrm{H}\hat{\mathbf{R}}^{-1}\mathbf{a}_0$ and $ 1/\bar{\mathbf{a}}_0^\mathrm{H}\hat{\mathbf{R}}^{-1}\bar{\mathbf{a}}_0$ can be considered as the power collected from the directions of the true SOI and the presumed SOI steering vector, respectively. Also, it is easy to show that $ 1/\mathbf{a}_0^\mathrm{H}\hat{\mathbf{R}}^{-1}\mathbf{a}_0$ can be approximated as the power of SOI-plus-noise. Since the Capon estimator has a good resolution in spectrum estimation, $ 1/\bar{\mathbf{a}}_0^\mathrm{H}\hat{\mathbf{R}}^{-1}\bar{\mathbf{a}}_0$ will deviate from $ 1/\mathbf{a}_0^\mathrm{H}\hat{\mathbf{R}}^{-1}\mathbf{a}_0 $ as long as mismatches occur between $\mathbf{a}_0$ and $\bar{\mathbf{a}}_0$ resulting in the distortion of the array
response. When large  errors occur, $ 1/\bar{\mathbf{a}}_0^\mathrm{H}\hat{\mathbf{R}}^{-1}\bar{\mathbf{a}}_0 $ is no longer equal to the SOI power but will be dominated by noise power. \cite{hao2019robust}.

\section{Interference DoAs and Power Spectral Estimation }\label{sec3}

Most of the existing INC matrix reconstruction methods are based on the Capon spatial power in \eqref{Power SMI}, which requires integration over $2 \pi$ angular sectors and many sampling points. Furthermore, they need prior information about the directions of the desired signal and interferences. To avoid the need for this prior information, and to decrease the number of sampling points and the computational complexity of numerically evaluating the integral, we present a technique to estimate the DoAs of the interferers over the available snapshots in which the angular sectors of the interfering signals are computed adaptively.
{For analysis purposes we use a setting with a single interferer, which is analytically tractable, to compare the power spectrum estimates obtained via the Capon spectrum and via maximum entropy.}  We investigate first the Capon spectrum estimation (section \ref{subsec3B}) and then, in section \ref{subsec3c}, the proposed maximum entropy power spectral estimation. 

\subsection{Estimation of the Interference DoAs }\label{subsec3a}

One of the most important and critical problems facing sensor array systems is the detection of the number of sources impinging on the array. This is a key step in most super-resolution estimation techniques, which often take the number of signals as prior information (It can be obtained through the minimum description length criterion in \cite{wax1989detection}).\\
\indent Estimation of the DoAs of the interference and corresponding uncertainty region during the time interval in which snapshots are taken is crucial for the implementation of the proposed CMR-ISPS method. In the proposed method a procedure is presented which enables the algorithm to adapt itself to the motion of interference by determining the angular sector where the null in the beampattern is to be created.\\
\indent {In this regard,} a DoA estimation technique using correlation is adopted where the inner products of the received vectors with the steering vector corresponding to a general direction of incidence are computed \cite{mohammadzadeh2019robust}. First, a coarse estimate of the DoAs is obtained from the discrete Fourier transform (DFT) of the first received vector in the set of snapshots 
$\mathbf{x}(1)=[x_1(1),\cdots,x_M(1)]^\mathrm{T}$ 
where $x_m(k)$ is the received signal at the \textit{m}-th antenna. Then, an angular sector centered on this estimate is scanned and the angle which maximizes the magnitude of the inner product is taken as the DoA estimate, as follows
\begin{align} \label{correlation estimator}
\theta_l(k)=\underset{\theta \in \Theta_\mathrm{x}}{\operatorname{argmax}}\ \arrowvert\mathbf{x}^H(k)\mathbf{a}(\theta)\arrowvert, \quad k=1,\cdots,K
\end{align}
where $ \Theta_\mathrm{x}= [\hat{\theta}_l- c, \hat{\theta}_l+ c] $ is the angular sector corresponding to the estimated  interference signal from the DFT process while $c\ll \pi $ is a small angle. It is noted that the parameter $c$ should be chosen large enough to guarantee that the correct direction is in the interval, but its exact value is not very important. In order to estimate the uncertainty region for each interferer, this procedure is then repeated for the next vector in the snapshots. When all snapshot vectors are processed, a polynomial of sufficiently high degree (in the proposed method, it is assumed to be a polynomial of degree 2) can be fitted to the set of DoA estimates, $\theta_{l_\mathrm{fit}}(k)$. Note that the angular range of variation of the interference DoA is determined using this polynomial. Computing the sector in which the interference DoA might vary as $\Xi_{ \theta_l}=\max(\theta_{l_\mathrm{fit}}(k))-\min(\theta_{l_\mathrm{fit}}(k)) $ {for $1 \le k \le K$}, then the 
width in the number of samples of this sector is defined by
\begin{align} \label{Theta_l}
{B_{\theta_l}}=\mathrm{floor}(\dfrac{\Xi _{\theta_l}}{2\pi /Q}),   
\end{align}
where $Q$ is defined as \eqref{denominator}, and floor(X) {truncates} each element of X to the nearest integer less than or equal to that element.
Aside from the correlation estimator (\ref{correlation estimator}), there are other methods that use low-resolution direction-finding \cite{huang2015robust,jidf,jidf_doa,jio,jio_doa,msesprit}.

\subsection{Capon Power Spectral Estimation}\label{subsec3B}
The most important issue with INC matrix reconstruction is the accuracy of the power spectrum estimate. Inaccuracies in the power spectral estimate result in distorted angular positions of interfering signals and their powers, which eventually lead to their insufficient suppression. In this subsection, we verify the influence of the interference and noise in {the power} Capon spectral estimation in \eqref{Power SMI} which will be utilized in subsection \ref{Analysis} for the INC matrix reconstruction analysis. 
To this end, we consider an analysis with a single interference that is amenable to a closed-form solution. Therefore,  
we assume that there is one interferer with  $\mathbf{a}(\theta_l)=\mathbf{a}_l$ and the SOI is not present in the training data \cite{zhu2020robust,zhu2019covariance}. Furthermore, we emphasize that the proposed algorithm is designed for scenarios with multiple interferes and only the analysis is demonstrated for a single interferer because of its simplicity and mathematical tractability. Hence, for this case, we can rewrite the theoretical covariance matrix in \eqref{Theoretical R}  as follows
\begin{align}\label{R capon}
\mathbf{R}_{tc}=\sigma^2_\mathrm{n}\mathbf{I}_M+\sigma^2_l\mathbf{a}_l\mathbf{a}_l^\mathrm{H}.
\end{align}
Based on the Capon spectrum estimator in \eqref{Power SMI}, the authors in \cite{gu2012robust} proposed an algorithm to utilize $\hat{P}_\text{(Cap)in}(\theta)$ as the power spectrum in the interference-plus-noise spatial domain 
\begin{align}\label{Pcapon}
 \hat{P}_\text{(Cap)in}(\theta)= {1} / {\mathbf{a}^\mathrm{H}(\theta)\mathbf{R}^{-1}_{tc}\mathbf{a}(\theta)}
 \end{align}
{It is assumed that $\theta \in \Theta_{\mathrm{in}}$ and $\Theta_{\mathrm{in}}$ is the angular sector of the interferences-plus-noise.}
{Then,} the inverse of the covariance matrix in (\ref{R capon}) 
by applying  Woodbury's matrix inversion lemma {is expressed} as 
\begin{align}\label{App Rc invers}
\mathbf{R}^{-1}_{tc}=\frac{1}{\sigma^2_\mathrm{n}} \Big(\mathbf{I}_M-\frac{\mathbf{a}_l\mathbf{a}_l^\mathrm{H}}{\rho+\|\mathbf{a}_l\|^2} \Big),
\end{align}
where $\| \mathbf{a}_l \|^2=M$ and $ \rho=\sigma^2_\mathrm{n}/\sigma^2_l $. In order to compute the power spectrum in the interference-plus-noise spatial domain and simplify the calculation, we assume that $\Theta_\mathrm{in}$ is sampled with a finite number of angles as $ \{ \theta_j\}_{j=1}^Q $, and the denominator of (\ref{Pcapon}) is re-written by substituting (\ref{App Rc invers}) as follows
\begin{align}\label{denominator}
\mathbf{a}^\mathrm{H}(\theta_j)\mathbf{R}_{tc}^{-1}\mathbf{a}(\theta_j)=\frac{1}{\sigma^2_\mathrm{n}}\Big(M-\frac{\arrowvert\mathbf{a}^H(\theta_j)\mathbf{a}_l\arrowvert^2}{\rho+M} \Big).
\end{align}
where $ \|\mathbf{a}_l\|^2=\|\mathbf{a}(\theta_j)\|^2=M$. Note that an analytical evaluation of (\ref{denominator}) may be difficult. A rough estimate is achieved when the inner product is approximated as
\begin{equation}\label{norm}
\arrowvert\mathbf{a}^\mathrm{H}(\theta_j)\mathbf{a}_l\arrowvert^2\simeq
\begin{cases}
M^2, & \theta_j = \theta_l
\\
0, & \theta_j \neq \theta_l
\end{cases}
\end{equation}
It is assumed that the approximation is correct if the angles $ \{ \theta_j\}_{j=1}^Q $, are chosen in such a way that only one of them coincides with the interference direction, $ \theta_l $ (that is, the other angles, $\theta_j \neq \theta_l$ fall outside the main-beam of the function $ \arrowvert\mathbf{a}^\mathrm{H}(\theta_j)\mathbf{a}_l\arrowvert^2 $) and, in order to have only one angle within the main beam of the interference power spectrum, the number of sensors, $ M $ should be large enough ($M > L $). By replacing \eqref{norm} into \eqref{denominator} and rearranging \eqref{Pcapon}, we obtain the approximated Capon power spectrum of the DoA of the interference $\theta_l$, as follows \vspace{-0.4em}
\begin{align} \label{Final Pcapon}
 \hat{P}_\text{(Cap)in}(\theta_{l}) \approx \sigma^2_l+\dfrac{\sigma^2_\mathrm{n}}{M}. 
\end{align}
\eqref{Final Pcapon} implies that in the impinging direction, the power is the sum of the residual noise and the interference signal, and the power of residual noise has become ${1} / {M}$ of actual noise.
\subsection{Maximum Entropy Power Spectrum Estimation} \label{subsec3c}

In the proposed CMR-ISPS beamforming method, an approach different from prior works is adopted to estimate the power spectrum of the signals utilizing the maximum entropy power spectral estimator \cite{MEPSalgorithm}. The essence of the idea is based on the use of the spatial spectrum distribution over all possible directions and 
estimates of the angular regions where the desired signal, {$\Theta_\mathrm{s}$} and the interferers, {$\Theta_{\mathrm{ME}_L}$} lie. 

One of the drawbacks of standard techniques {(i.e. Capon beamformer)} to spectrum estimation is that the autocorrelation sequence can only be estimated for displacements of $ \arrowvert k \arrowvert < M $ in the observation vector of length $ M $. Thus, it is set to zero for $ \arrowvert k\arrowvert \geq M $. However, this windowing may significantly limit the accuracy of the estimated spectrum and resolution since many signals of interest have autocorrelation that is nonzero for $ \arrowvert k\arrowvert \geq M $. This is especially true for narrowband processes with an autocorrelation that decays slowly with $t$. However, we demonstrate that the maximum entropy method does not make this assumption, and extrapolates the autocorrelation sequence in such a way that the entropy of the spectrum is minimized. This leads to a more accurate representation of the power spectrum by the available autocorrelation matrix estimate and a more accurate reconstruction of the INC. Inaccuracies in the power spectral estimate result in distorted angular positions of interference signals, as well as their powers, which eventually lead to their insufficient suppression. Thus, the maximum entropy power spectrum estimation is a crucial step in enhancing the SINR performance of beamformers. 

Now, it is supposed that we are given the first $M$ values $r_{\mathrm{x},0}, r_{\mathrm{x},1}, \cdots, r_{\mathrm{x},(M-1)}$ of the autocorrelation sequence and we wish to find a model, by choosing the remaining values $ \arrowvert k\arrowvert > M $ where the autocorrelation is given as $r_{\mathrm{x},M}=\mathrm{E} \big \{x(t+M)  x^*(t) \big\} $. An estimate of $r_{\mathrm{x},M}$ lies outside the interval of known autocorrelation function values. Basic autocorrelation 
{properties }state that $r_{\mathrm{x},M}$ must have a value such that the $(M + 1) \times (M + 1)$ Hermitian Toeplitz autocorrelation matrix given by \cite{cabanes2019toeplitz}
\begin{equation}\label{matrix}
\mathbf{R} \triangleq
\begin{bmatrix}
r_{x,0}&r_{x,-1}&\cdots& r_{x,-M}\\
r_{x,1}& r_{x,0}&\cdots& r_{x,(1-M)}\\
\vdots&\vdots&     &\vdots&\\
r_{x,M}& r_{x,(M-1)}& \cdots & r_{x,0}
\end{bmatrix},
\end{equation}
is positive semidefinite (i.e., all subdeterminants of $\mathbf{R}$ must be nonnegative). Since $\det(\mathbf{R})$ is a quadratic function of $r_{\mathrm{x},M}$, two values of $r_{\mathrm{x},M}$ make the determinant equal to zero. These two values of $r_{\mathrm{x},M}$ define boundaries within which the predicted value of $r_{\mathrm{x},M}$ must fall. The maximum entropy method procedure seeks to select the value of $r_{\mathrm{x},M}$ that maximizes $\det(\mathbf{R})$. The most important application of maximum entropy power spectral estimation has been in estimating a power spectrum from partial knowledge of its autocorrelation function, $\{r_{\mathrm{x},k}\}_{-M\leq k \leq M }$. 
In this situation, the power spectrum based on maximum entropy is estimated as \cite{MEPSalgorithm}
\begin{align}\label{Power of MEM}
\hat{P}_\mathrm{ME}(\theta)=\dfrac{1}{\epsilon\big(\mathbf{a}^\mathrm{H}(\theta)\hat{\mathbf{R}}^{-1} \mathbf{u}_1 \mathbf{u}^\mathrm{H}_1\hat{\mathbf{R}}^\mathrm{-H}\mathbf{a}(\theta)\big)}  =\dfrac{1}{\epsilon|\mathbf{a}^\mathrm{H}(\theta)\hat{\mathbf{R}}^{-1} \mathbf{u}_1|^2},
\end{align}
where \add{$\mathbf{u}_1=[\begin{smallmatrix}1, & 0, & \cdots, & 0\end{smallmatrix}]^T$} and $\epsilon=1/\mathbf{u}_1^\mathrm{T}\hat{\mathbf{{R}}}^{-1}\mathbf{u}_1$. 
To compare the power spectrum of the maximum entropy {\eqref{Power of MEM}} with the Capon estimator in \eqref{Final Pcapon}, we assume that the theoretical covariance matrix can be expressed as \eqref{R capon} and we choose a finite number of angles $\theta_j \in \Theta_\mathrm{in} (j=1,2,\cdots,Q)$ to discretize the angular sector $\Theta_\mathrm{in}$. Hence, \eqref{Power of MEM} becomes
\begin{align}\label{Power of MEM disc.}
\hat{P}_\mathrm{ME}(\theta_j)&=\dfrac{1}{\epsilon\big(\mathbf{a}^\mathrm{H}(\theta_j)\mathbf{R}_{tc}^{-1} \mathbf{u}_1 \mathbf{u}^\mathrm{H}_1\mathbf{R}_{tc}^\mathrm{-H}\mathbf{a}(\theta_j)\big)} 
\end{align}
where $\mathbf{R}_{tc}^\mathrm{-1}$ is given by \eqref{App Rc invers}.
By replacing \eqref{App Rc invers} into \eqref{Power of MEM disc.}, and assuming that one and only one sample of the discrete sample points $\theta_j$ lies in the main beam and coincides with the direction of the interference $\theta_l$, {if we choose $\theta_j=\theta_l$,} the denominator of \eqref{Power of MEM disc.} can be written as \eqref{Deno_MEM}. 
\begin{figure*}[t] 
	\begin{align}\label{Deno_MEM}
	\epsilon \Big(\mathbf{a}^\mathrm{H}(\theta_l)\hat{\mathbf{R}}^{-1} \mathbf{u}_1 \mathbf{u}^\mathrm{H}_1\hat{\mathbf{R}}^\mathrm{-H}\mathbf{a}(\theta_l)\Big)&=\epsilon \Big( \mathbf{a}^\mathrm{H}(\theta_l)\bigg[ \frac{1}{\sigma^2_\text{n}} \Big(\mathbf{I}_M-\frac{\mathbf{a}_l\mathbf{a}_l^H}{\rho+\|\mathbf{a}_l\|^2} \Big)\bigg]\mathbf{u}_1 \mathbf{u}^\mathrm{H}_1 \bigg[ \frac{1}{\sigma^2_\text{n}} \Big(\mathbf{I}_M-\frac{\mathbf{a}_l\mathbf{a}_l^H}{\rho+\|\mathbf{a}_l\|^2} \Big)\bigg] \mathbf{a}(\theta_l) \Big) \nonumber \\ &= \epsilon \bigg[ \frac{1}{\sigma^2_\text{n}} \Big(\mathbf{a}^\mathrm{H}(\theta_l)\mathbf{u}_1-\frac{\mathbf{a}^\mathrm{H}(\theta_l) \mathbf{a}_l\mathbf{a}_l^H \mathbf{u}_1}{\rho+\|\mathbf{a}_l\|^2} \Big)\bigg] \bigg[ \frac{1}{\sigma^2_\text{n}} \Big(\mathbf{u}^\mathrm{H}_1\mathbf{a}(\theta_l)-\frac{ \mathbf{u}^\mathrm{H}_1 \mathbf{a}_l\mathbf{a}_l^H \mathbf{a}(\theta_l)}{\rho+\|\mathbf{a}_l\|^2} \Big)\bigg]  \nonumber \\
    &= \epsilon \bigg[ \frac{1}{\sigma^2_\text{n}} \Big(1-\frac{M}{\rho+M} \Big)\bigg] \bigg[ \frac{1}{\sigma^2_\text{n}} \Big(1-\frac{M}{\rho+M} \Big)\bigg]=  \frac{\epsilon}{\sigma^4_\text{n}} \Big( \frac{\rho}{\rho+M} \Big)^2 =\epsilon \Big( \frac{1}{\sigma^2_\text{n}+\sigma^2_l M} \Big)^2
	\end{align}
 	\noindent\rule{\textwidth}{.5pt}
\end{figure*}
Based on the above explanation, the maximum entropy power spectrum estimation of the interference $\theta_l$ can be written as
\begin{align} \label{Phat with epsilon}
 \hat{P}_\mathrm{ME}(\theta_l) \approx { ( \sigma^2_\text{n}+\sigma^2_l M )^2} / {\epsilon}
\end{align}
where 
\begin{equation} \label{epsilon}
 \epsilon =\dfrac{1}{\mathbf{u}_1^\mathrm{T}\mathbf{R}_{tc}^{-1}\mathbf{u}_1}= \dfrac{\sigma^2_\text{n}(\sigma^2_\text{n}+M \sigma^2_l)}{\sigma^2_\text{n}+(M-1) \sigma^2_l} \approx \sigma^2_\text{n},  \quad (M\gg 1)
\end{equation}
By replacing \eqref{epsilon} into \eqref{Phat with epsilon} 
\begin{align} \label{final Phat mem}
 \hat{P}_\mathrm{ME}(\theta_l)&= \dfrac{ ( \sigma^2_\text{n}+\sigma^2_l M )^2}{\sigma^2_\text{n}}= \sigma^2_\text{n}+(2M+M^2  \text{INR})\sigma^2_l
\end{align}
Also from \eqref{Final Pcapon}, we can derive that
\begin{align} \label{Pcapon with INR}
 \hat{P}_\text{(Cap)in}(\theta_l) = \dfrac{1}{M}\sigma^2_\text{n}+\sigma^2_l= \sigma^2_\text{n}(\dfrac{1}{M}+\text{INR})
\end{align}
where $\text{INR}=\sigma^2_l/\sigma^2_\text{n}$.
Comparing \eqref{final Phat mem} and \eqref{Pcapon with INR}, the presence of the {strong} interference power in the maximum entropy power spectrum is striking.
\begin{figure}[h]
	\centering
	\includegraphics[width=2.8in,height=2.1in]{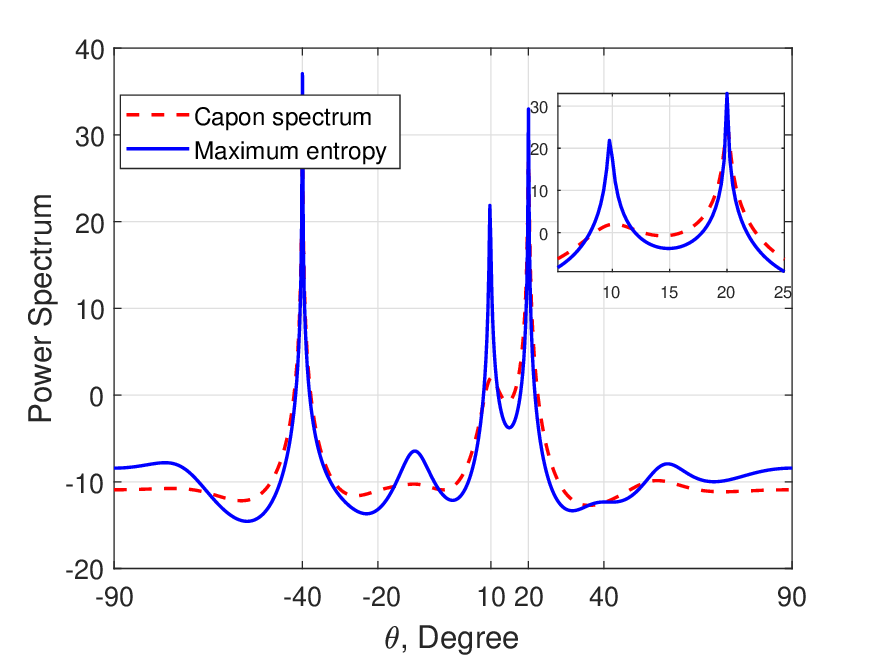}
	\caption{Power spectrum of maximum entropy versus Capon. }
	\label{Power spectrum}
\end{figure} 
{Therefore}, as it is shown in Fig.~\ref{Power spectrum}, the maximum entropy has more pronounced peaks  
around the DoA of the interference and correspondingly the interferences contribute more to the INC matrix reconstruction.

Fig.~\ref{Power spectrum} represents the power spectrum curves {obtained using} the Capon spectral estimation, \eqref{Pcapon} and maximum entropy power spectrum, \eqref{Power of MEM}. Assume that the SOI and two interference signals impinge on the half-spaced uniform linear array with $M=10$ from directions $\bar{\theta}_0=10^\circ$, $\theta_1=20^\circ$, and $\theta_2=-40^\circ$, respectively. In this simulation, the SNR is fixed at 10 dB and INR=30 dB and additive Gaussian noise have a unit variance. The curves corresponding to $\hat{P}_\mathrm{ME}(\theta)$ and $\hat{P}_\text{(Cap)in}(\theta)$ are the power spectrum of \eqref{SCM} which is the sample covariance matrix. Although the DoA of the SOI and interference are close, the high resolution of the maximum entropy power spectrum estimation is effective for the SOI and interference directions. From Fig.~\ref{Power spectrum}, it is evident that the Capon spectrum estimator has a wider mainlobe with respect to both desired and interference signals. Because of this, the accuracy of the Capon spatial spectrum degrades when coherent signals (with line spectra) exist \cite{wang2016robust}.

\section{Interference-plus-noise Covariance Matrix}\label{sec4}
In what follows, we detail an algorithm to reconstruct the INC  matrix using a weighted sum of the outer products of steering vectors, the coefficients of which can be estimated in the vicinity of the DoAs of the interferences as follows
\begin{align}\label{C}
\hat{\mathbf{R}}_{\mathrm{ME}}=\int_{\Theta_\mathrm{in}} \hat{P}_\mathrm{ME}(\theta) \mathbf{{a}}(\theta)\mathbf{{a}}^H(\theta)d\theta 
\end{align} 
where $ \mathbf{{a}}(\theta) $ is the  associated with a hypothetical direction $ (\theta) $ based on the known array structure and $ \hat{P}_\mathrm{ME}(\theta) $ may be interpreted as the spatial power spectrum of the interference-plus-noise component of the received signal. 
The integral can be approximately calculated using a discrete sum 
\begin{align}\label{C with sigma}
\hat{\mathbf{R}}_{\mathrm{ME}}\cong\sum_{j=1}^{Q}\hat{P}_\mathrm{ME}(\theta_j) \mathbf{{a}}(\theta_j)\mathbf{{a}}^H(\theta_j) \Delta \theta_j \vspace{5mm} \ \ \mathrm{for} \ \ Q\gg M
\end{align}
where $\Delta\theta_j \approx \frac{2\pi}{M}$. The cost of INC matrix reconstruction with  \eqref{C with sigma} requires a large number of computations to be able to accurately synthesize powers from signals for the whole $2 \pi$ sector. In order to reduce complexity, we propose to sample only the sectors next to the interference and the SOI. Therefore, we assume that the DoAs of all interferences
always lie in small angular sectors $\Theta_l, l=1,2,\cdots, L$  such that the $l$th interference is located {in $\Theta_l$}, $\Theta_l \subset \Theta_{\text{in}}$ and ${\text{length}} \Theta_l \ll {\text{length}} \Theta_{\text{in}}$. (
Assuming that the SOI is located in the angular sector $ \Theta_\text{s} $, which can be obtained using a low-resolution direction finding method \add{\cite{schell199318}}, $\Theta_\mathrm{in}$ is a region containing the interference plus noise such that $ \Theta_{\text{in}} \cup \Theta_\text{s}  = [-\pi, \pi]$ and $ \Theta_{\text{in}} \cap \Theta_\text{s}  = \emptyset$).\\
\indent Thereby, a simplified approach is adopted where the interference angular sector is sampled for a short interval. This is in contrast to other methods in which a large sector $[-\pi, \pi]$ is considered to find the interference directions. To this end, we assume that $ L $ is the number of interferences to be suppressed, and the range $(b_l\leq b \leq b_l+{B_{\theta_l}}-1)$ corresponds to an angular sector centered on the DoA of the $l$th interference and the parameter $ {B_{\theta_l}} $ given in \eqref{Theta_l} determines the width of this range while the index of the first angular interval of this range, $b_l$ can be calculated as
\begin{align} \label{ni}
b_l=\mathrm{floor}(N_{lc}-\dfrac{1}{2}{B_{\theta_l}}),
\end{align}
where $  N_{lc}=\mathrm{ceil}(\frac{\theta_{lc}}{2\pi /Q}) $ is the number of angular {samples} corresponding to the center of the range, and assuming that the center of this range is $ \theta_{lc}=\big(\max(\theta_{l_\mathrm{fit}}(k))+\min(\theta_{l_\mathrm{fit}}(k) \big))/2 $. Note that, ceil(X) rounds each element of X to the nearest integer greater than or equal to that element. With $Q_{\text{ME}_L}=\bigcup \limits_{l=1}^{L}\{b_l\leq b \leq b_l+{B_{\theta_l}}-1 \}$ as the union of the small angular sectors of the interferences, we can rewrite the maximum entropy INC matrix {in \eqref{C with sigma}} exploiting the new interval as {a new INC matrix, $\hat{\mathbf{R}}_{\mathrm{ME}_L}$ as} follows
\begin{align}\label{proposed Rin summation}
\hat{\mathbf{R}}_{\mathrm{ME}_L} \cong \sum_{\theta_j \in \Theta_{\text{ME}_L}} \dfrac{\mathbf{a}(\theta_j)\mathbf{a}^\mathrm{H}(\theta_j)}{\epsilon|\mathbf{a}^\mathrm{H}(\theta_j)\hat{\mathbf{R}}^{-1} \mathbf{u}_1|^2}\Delta\theta_j 
\end{align}
where $\Theta_{\mathrm{ME}_L}= \bigcup _{l=1}^L \Theta_l$. It should be noted that $Q_{\text{ME}_L}\ll Q$, is the discretized number of sampling points for $\Theta_{\mathrm{ME}_L}$. Throughout the above analysis, the INC matrix of the proposed method is summed over a much smaller angular sector $\Theta_{\mathrm{ME}_L}$, which contains the actual interference DoAs as far as possible to improve the INC matrix reconstruction. 

\section{Estimation of the desired signal steering vector}\label{sec5}
In practice, it is difficult to obtain the actual steering vector by using the presumed DoA of the signal because of the propagation environment. Therefore, in this subsection, we describe a simple method in which the actual steering vector is estimated based on the SOI covariance matrix. 
The SOI covariance matrix can then be reconstructed based on the maximum entropy power spectrum by evaluating it over $\Theta_\text{s}$ as follows
\begin{align}\label{proposed Rs}
\hat{\mathbf{R}}_\mathrm{s} \approx \sum_{i=1}^{S}
\hat{P}_\mathrm{ME}(\theta_{s_i})  \mathbf{a}(\theta_{s_i})\mathbf{a}^\mathrm{H}(\theta_{s_i})\Delta\theta_s,
\end{align} 
where $S$ is the number of sampling points in the SOI angular sector spaced by $\Delta\theta_s$ {, and we choose a finite number of angles $\theta_{s_i} \in \Theta_\mathrm{s} (i=1,2,\cdots,S)$ to discretize the angular sector $\Theta_\mathrm{s}$}. 
The estimated SOI covariance matrix $ \hat{\mathbf{R}}_\mathrm{s} $, contains residual components arising from the white noise {as well as the SOI component, which}  can be expressed as
\begin{align} \label{estmated Rs}
    \hat{\mathbf{R}}_\mathrm{s} \cong \sigma^2_{\text{n}(\text{res})} \mathbf{Z}+ \hat{\sigma}^2_0 \mathbf{a}_0 \mathbf{a}_0^\mathrm{H},
\end{align}
where $ \hat{\sigma}^2_0 $ is the estimate of the SOI power, $ \sigma^2_{\text{n}(\text{res})} $ is the residual noise power (which is inside the estimated desired signal covariance matrix in \eqref{estmated Rs}) and $ \mathbf{a}_0 $ stands for the SOI steering vector {and $\| \mathbf{Z}\|_2=1$}. Multiplying the reconstructed SOI covariance matrix by the SOI presumed  $\bar{\mathbf{a}}_0$, we obtain the following estimate of the steering vector for the SOI
\begin{align}\label{ahat}
\hat{\mathbf{a}}_0=\hat{\mathbf{R}}_\mathrm{s} \bar{\mathbf{a}}_0=\sigma^2_{\text{n}(\text{res})} \mathbf{Z} \bar{\mathbf{a}}_0+\hat{\sigma}^{2}_0(\mathbf{a}^\mathrm{H}_0 \bar{\mathbf{a}}_0)\mathbf{a}_0.
\end{align}
In \eqref{ahat} the residual term $\sigma^2_{\text{n}(\text{res})} \mathbf{I}$  represents the noise power that falls within the SOI angular sector.
Since the SOI angular sector is much smaller than the full set of angles, $2 \pi$, the noise power in this sector is much smaller than the total noise power. Hence, the norm of the residual term  $\sigma^2_{\text{n}(\text{res})} \| \bar{\mathbf{a}}_0 \|$  can be expected to be much smaller than the power of the signal term $\hat{\sigma}^2_0 | \mathbf{a}_0^\mathrm{H} \bar{\mathbf{a}}_0 |$ {($ \|\sigma^2_{\text{n}(\text{res})} \mathbf{Z} \bar{\mathbf{a}}_0\| \le  \sigma^2_{\text{n}(\text{res})} \| \bar{\mathbf{a}}_0\| $)}. This can be better understood by noting that $\| \bar{\mathbf{a}}_0 \|^2= \| \mathbf{
a}_0 \|^2= M$  for the ideal form of the , and $|\mathbf{a}_0^\mathrm{H} \bar{\mathbf{a}}_0 | \approx M$  if $\bar{\mathbf{a}}_0$ is close to $\mathbf{a}_0$. Then it is sufficient that $\sigma^2_{\text{n}(\text{res})} \ll M \hat{\sigma}^2_0 $, which can be satisfied even for low SNR values. The accuracy of the steering vector estimate \eqref{ahat} can be investigated by calculating the beamformer SINR as follows.

In the derivation of the SINR for the beamformer based on the steering vector estimate in \eqref{ahat}, we assume that the INC matrix is exact (i.e. $\hat{\mathbf{R}}_{\text{in}}=\mathbf{R}_{\text{in}}$ ). This may be justified by noting that the exclusion of the SOI angular sector in the reconstruction of $\mathbf{R}_{\text{in}}$ is negligible if this sector is much smaller than $2\pi$. Then, the SINR becomes
\begin{align} \label{sinr as}
\mathrm{SINR}=\sigma^2_0 \dfrac{| \hat{\mathbf{a}}_0^\mathrm{H} {\mathbf{R}}_{\text{in}}^{-1}{\mathbf{a}}_0 |^2}{\hat{\mathbf{a}}_0^\mathrm{H} {\mathbf{R}}_{\text{in}}^{-1}\hat{\mathbf{a}}_0}.
\end{align}
when $\hat{\mathbf{a}}_0= \mathbf{a}_0$, the optimum SINR is given by
\begin{align}\label{eq:optimal.SINR}
\mathrm{SINR}_\text{opt}=\sigma^2_0 | \mathbf{a}_0^\mathrm{H} \mathbf{R}_{\text{in}}^{-1}{\mathbf{a}}_0 |.
\end{align}
By direct substitution of \eqref{ahat} into \eqref{sinr as} and using the approximation $(1+x)^{-1} \approx 1-x$ for $|x| \ll 1$, we can write
\begin{align} \label{SINR as}
\mathrm{SINR}=&\sigma^2_0 | \mathbf{a}_0^\mathrm{H} \mathbf{R}_{\text{in}}^{-1}{\mathbf{a}}_0 | \Big \lbrace 1- \epsilon^2_\text{res} \nonumber \\ & \cdot \Big [ \dfrac{|\bar{\mathbf{a}}_0^\mathrm{H} \mathbf{R}_{\text{in}}^{-1}\bar{\mathbf{a}}_0| | \mathbf{a}_0^\mathrm{H} \mathbf{R}_{\text{in}}^{-1} \mathbf{a}_0|- | \bar{\mathbf{a}}^\mathrm{H} \mathbf{R}_{\text{in}}^{-1} \mathbf{a}_0|^2}{|\bar{\mathbf{a}}_0^\mathrm{H} \mathbf{a}_0|^2 | \mathbf{a}_0^\mathrm{H} \mathbf{R}_{\text{in}}^{-1} \mathbf{a}_0|^2} \Big] \Big \rbrace
\end{align}
where $\epsilon_\text{res}=\frac{\sigma^2_{\text{n}(\text{res})}}{\hat{\sigma}_0^2}$ is much less than one. Now, an insight into the dependence of the reduction in SINR on the error in the presumed  $\bar{\mathbf{a}}_0$ can be obtained by considering the single interference case. For this case, we exploit the INC matrix defined in \eqref{R capon} where $\mathbf{R}_{\text{in}}= \sigma_l^2\mathbf{a}_l\mathbf{a}_l^\mathrm{H}+\sigma_\text{n}^2 \mathbf{I}$. The inverse of the INC matrix can be obtained as in \eqref{App Rc invers}. Therefore, with this inversion, the numerator of the expression within the square brackets in \eqref{SINR as} becomes
\begin{align} \label{long formula}
    |\bar{\mathbf{a}}_0^\mathrm{H} \mathbf{R}_{\text{in}}^{-1}\bar{\mathbf{a}}_0| | \mathbf{a}_0^\mathrm{H} &\mathbf{R}_{\text{in}}^{-1} \mathbf{a}_0|- | \bar{\mathbf{a}}_0^\mathrm{H} \mathbf{R}_{\text{in}}^{-1} \mathbf{a}_0|^2= \dfrac{1}{\sigma^4_\text{n}} \Big(M^2- |\bar{\mathbf{a}}_0^\mathrm{H} \mathbf{a}_0|\nonumber \\&+\dfrac{2}{M+\rho} \Re \Big \lbrace(\bar{\mathbf{a}}_0^\mathrm{H}\mathbf{a}_0) (\mathbf{a}_0^\mathrm{H} \mathbf{a}_l) (\mathbf{a}_l^H \bar{\mathbf{a}}_0) \Big \rbrace \nonumber \\ & - \dfrac{M}{M+\rho}(|\bar{\mathbf{a}}_0^\mathrm{H} \mathbf{a}_l |^2+ |\mathbf{a}^\mathrm{H}_0 \mathbf{a}_l|^2) \Big )
\end{align}
where $\Re \lbrace \cdot \rbrace$ denotes the real part of a complex number. Then, it can be shown that \begin{align}
    \bar{\mathbf{a}}_0^\mathrm{H} \mathbf{a}_0=&\sum_{q=0}^{M-1} e^{-j q (\theta_0-\bar{\theta}_0)}=\dfrac{1-e^{jM(\bar{\theta}_0-\theta_0)}}{1-e^{j(\bar{\theta}_0-\theta_0)}} \nonumber \\ =&\dfrac{e^{j\frac{M}{2}(\bar{\theta}_0-\theta_0)}\Big(2j \sin (\frac{M}{2}(\bar{\theta}_0-\theta_0)) \Big ) }{e^{j\frac{1}{2}(\bar{\theta}_0-\theta_0)}\Big(2j \sin (\frac{1}{2}(\bar{\theta}_0-\theta_0)) \Big )}
\end{align}
and then we can write
\begin{align} \label{sinc}
  |\bar{\mathbf{a}}_0^\mathrm{H} \mathbf{a}_0|^2= \dfrac{\sin^2 \big(\frac{M}{2}(\bar{\theta}_0-\theta_0)\big )}{\sin^2 \big(\frac{1}{2}(\bar{\theta}_0-\theta_0)\big)}
\end{align}
where $\bar{\theta}_0$ is the DoA corresponding to the presumed steering vector, $\bar{\mathbf{a}}_0$. Also, if the interference DoA is sufficiently separated from the DoAs of $\bar{\mathbf{a}}_0$ and $\mathbf{a}_0$, the contributions of the terms in \eqref{long formula} involving the interference steering vector $\mathbf{a}_l$ become negligible compared to the first two terms. Using the following Taylor series expansion of the right-hand side of \eqref{sinc} with $\Phi=\bar{\theta}_0-\theta_0$, assumed to be much smaller than one,
\begin{align} 
   \dfrac{\sin^2(M \Phi/2)}{\sin^2(\Phi/2)} \approx M^2- \dfrac{1}{4}M^2(M^2-1)\Phi^2
\end{align}
then we obtain
\begin{align}\label{eq:approx.SINR}
    \mathrm{SINR} \approx \sigma^2_0 | \mathbf{a}_0^\mathrm{H} \mathbf{R}_{\text{in}}^{-1}{\mathbf{a}}_0 | \big( 1- \dfrac{1}{12} \epsilon^2_\text{res} \Phi^2 \big )
\end{align}
The residual noise power in the SOI angular sector may be taken to be proportional to the width of this sector, assuming that the noise is spatially white. Then, the residual noise power can be expressed as
\begin{align}
    \sigma^2_{\text{n}(\text{res})}= {S \sigma^2_\text{n}} / {Q}
\end{align}
Hence, we have
\begin{align}
    \epsilon_\text{res}=\dfrac{\sigma^2_{\text{n}(\text{res})}}{\hat{\sigma}^2_0} \approx \dfrac{S}{Q \times \text{SNR}}
\end{align}
Since, $S \ll Q$, $\epsilon_\text{res}$ is much smaller than one even for low SNR values. Comparing \eqref{eq:optimal.SINR} with \eqref{eq:approx.SINR}, we notice that the vector $ \hat{\mathbf{a}}_0 $ is a good estimate for the SOI steering vector.

\section{Iterative maximum entropy beamformer }\label{sec6}
{Thus far, the DoA of the signals and corresponding angular sectors, the INC matrix, and the desired signal  have been computed. Now, by inserting the INC matrix, \eqref{proposed Rin summation} and the desired signal, \eqref{ahat} into \eqref{optimal wegight vector}, the weight vector could be calculated as follows: }
\begin{align}\label{Pro1 wegight vector}
\mathbf{w}_{\mathrm{pro}}=\dfrac{\hat{\mathbf{R}}_{\mathrm{ME}_L}^{-1} \hat{\mathbf{a}}_0}{\hat{\mathbf{a}}_0^\mathrm{H} \hat{\mathbf{R}}_{\mathrm{ME}_L}^{-1}\hat{\mathbf{a}}_0},
\end{align}
However, it is well-known that the computational complexity for the inverse of $M \times M$ matrix is $\mathcal{O}(M^{3})$. On the other hand, the INC matrix reconstruction in \eqref{proposed Rin summation} requires a complexity of $\mathcal{O}(Q_{\text{ME}_L}M^2)$ to synthesize the narrowband signal power accurately, where  $Q_{\text{ME}_L}\ll Q$. Therefore, in order to decrease the computational complexity and avoid the inverse implementation, we develop an efficient adaptive version of the maximum entropy power spectrum interference-plus-noise covariance technique based on the conjugate gradient method. The proposed CMR-ISPS algorithm updates the beamforming weights with a reduced cost as it does not explicitly form the covariance matrices, relying instead on low-cost iterative techniques. The estimated weight vector is obtained from a coarse estimate of the angular sector where the SOI lies using conjugate gradient iterations that avoid the explicit inversion of the covariance matrix. To the best of the authors' knowledge, the iterative conjugate gradient has not been investigated to decrease the computational complexity of matrix-vector multiplication and the INC matrix reconstruction in robust adaptive beamforming in the literature. Therefore, based on the following procedure, we show that the proposed CMR-ISPS needs low complexity {to compute the weight vector while it avoids the explicit inversion of the matrix.}

We apply a conjugate gradient approach \cite{frost1972algorithm,sjidf} to solve the MVDR optimization problem in (\ref{MVDR}) by using the Lagrange multiplier $\alpha$ to include the constraint into the objective function as \vspace{-0.4em}
\begin{align} \label{cost function}
 \text{min} \  J(\mathbf{w})=\mathbf{w}^\mathrm{H} \hat{\mathbf{R}}_{\mathrm{ME}_L} \ \mathbf{w}+\alpha(\mathbf{w}^\mathrm{H} \hat{\mathbf{a}}_0-1).
\end{align}
By solving this optimization problem, it is clear that multiplier $\alpha$ is solved in such that the constraint equation in \eqref{MVDR} is satisfied and $\alpha= 1 / \hat{\mathbf{a}}_0^\mathrm{H} \hat{\mathbf{R}}_{\mathrm{ME}}^{-1} \hat{\mathbf{a}}_0$. However, since the SINR of the beamformer is independent of $\alpha$, and for convenience, let $\mathbf{w}$ denote the current beamformer in the steepest descent algorithm. We have:\vspace{-0.7em}
\begin{align} \label{cost function f}
  \underset{{\mathbf{w}}}{\operatorname{min}} \lbrace f(\mathbf{w})=\dfrac{1}{2} \mathbf{w}^\mathrm{H} \hat{\mathbf{R}}_{\mathrm{ME}_L} \ \mathbf{w}+\mathbf{w}^\mathrm{H} \hat{\mathbf{a}}_0 \rbrace,
\end{align}
and let $\mathbf{d}$ denote the current direction. Starting from an initial search direction $\mathbf{d}_0$, the steepest descent method for quadratic problem generates search directions $\mathbf{d}_1, \mathbf{d}_2, \cdots, \mathbf{d}_{T_{\mathrm{it}}-1}$. 
Now let us compute the next iteration of the steepest descent algorithm using line search. If $\mu$ is the generic step-size, then
\begin{align}
    f(\mathbf{w}_t+\mu \mathbf{d}_t)=\dfrac{1}{2} (\mathbf{w}_t+&\mu \mathbf{d}_t)^\mathrm{H} \hat{\mathbf{R}}_{\mathrm{ME}_L}  (\mathbf{w}_t+\mu \mathbf{d}_t) \nonumber \\ &+(\mathbf{w}_t+\mu \mathbf{d}_t)^\mathrm{H} \hat{\mathbf{a}}_0.
\end{align}
By solving $\underset{{\mathbf{w}}}{\operatorname{min}} f(\mathbf{w}_t+\mu \mathbf{d}_t)=\frac{\partial f(\mathbf{w}_t+\mu \mathbf{d}_t)}{\partial \mu}=0$, the step size obtained is 
\begin{align} \label{mu_k}
    \mu_t=-{\mathbf{d}_t^\mathrm{T} \mathbf{g}_t } / {\mathbf{d}_t^\mathrm{T}\hat{\mathbf{R}}_{\mathrm{ME}_L} \mathbf{d}_t },
\end{align}
where $\mathbf{g}_t= \nabla  f(\mathbf{w}_t)$, and 
$\nabla  f(\mathbf{w}_t)$ is the gradient of the cost function with respect to $\mathbf{w}_t$ and $t=1,\cdots,T_{\mathrm{it}}$ is the iteration number. Moreover, the gradient vector obtained from (\ref{cost function f}) is
\begin{align} \label{Gradient}
  \nabla  f(\mathbf{w}_k)= \hat{\mathbf{R}}_{\mathrm{ME}_L} \mathbf{w}_k+ \hat{\mathbf{a}}_0. 
\end{align}
Here, we show how to update \eqref{Gradient} while avoiding computing \eqref{proposed Rin summation} explicitly with the complexity of $\mathcal{O}(Q_{\text{ME}_L}M^2)$.
Rewriting (\ref{Gradient}) by substituting the expression for $\hat{\mathbf{R}}_{\mathrm{ME}_L}$, we get
\begin{equation} \label{Grad with w}
 \nabla  f(\mathbf{w}_t) = 
  \sum_{i=1}^{Q_{\text{ME}_L}} \hat{P}_\mathrm{ME}(\theta_i) \Big(\mathbf{a}^\mathrm{H}(\theta_i)   \mathbf{w}_t\Big)\mathbf{a}(\theta_i)\Delta\theta + \hat{\mathbf{a}}_0, \end{equation}
It is evident that the complexity for the above computation is $\mathcal{O}(Q_{\text{ME}_L}M)$.
{Then}, by substituting (\ref{Grad with w}) instead of $\mathbf{g}_k$ and updating \eqref{mu_k} in every iteration, {$k$}, the cost function $f(\mathbf{w})$  can be minimized by applying the conjugate gradient algorithm and the beamforming weight vector is updated at each iteration as follows 
\begin{align} \label{final W}
  \mathbf{w}_{t+1}=\mathbf{w}_t + \mu_t \mathbf{d}_t, \quad (t\ge 0)
\end{align}
where $\mathbf{w}_t$ is the current iteration, $\mathbf{d}_t$ is a descent direction of $f$ at $\mathbf{w}_t$ and $\mu_t$ is the step-size. The search direction $\mathbf{d}_t$ is guaranteed to have a descent direction due to $\mathbf{g}_t^\mathrm{T} \mathbf{d}_t \le 0$. The directions $\mathbf{d}_t$ are generated in the light of classical conjugate direction methods \cite{hestenes1952methods,polyak1969conjugate} as
\begin{align} \label{Power Method}
\mathbf{d}_k=
    \begin{cases}
    -\mathbf{g}_t, \qquad \qquad   \ \ t=0 \\ 
     -\mathbf{g}_t+ \beta_t \mathbf{g}_{t-1}, \ \  t \geq 1
    \end{cases}       
\end{align}
where $\beta_t \mathbf{g}_{t-1}$ is the correction term for the steepest descent direction $-\mathbf{g}_t$, and $\beta_t$ is  presented as 
\begin{align}
    \beta_t= {\mathbf{g}_{t}^\mathrm{T} (\mathbf{g}_{t}-\mathbf{g}_{t-1})} / {  \parallel \mathbf{g}_{t-1}  \parallel^2}.
\end{align}
Some well-known conjugate gradient methods are available, such as FR (Fletcher–Revees) \cite{fletcher1964function}, PRP (Polak–Ribière–Polyak) \cite{beamcg}, and HS (Hestenes–Stiefel)  conjugate gradient method, respectively. Among these, the PRP method is considered the best in practical computation \cite{beamcg}. To compute the CMR-ISPS beamformer, the proposed extension of the conjugate gradient algorithm is used to solve the optimization problem in \eqref{cost function} as algorithm 1.

\begin{figure}[h]
	\centering
		\includegraphics[width=2.7in,height=2.1in]{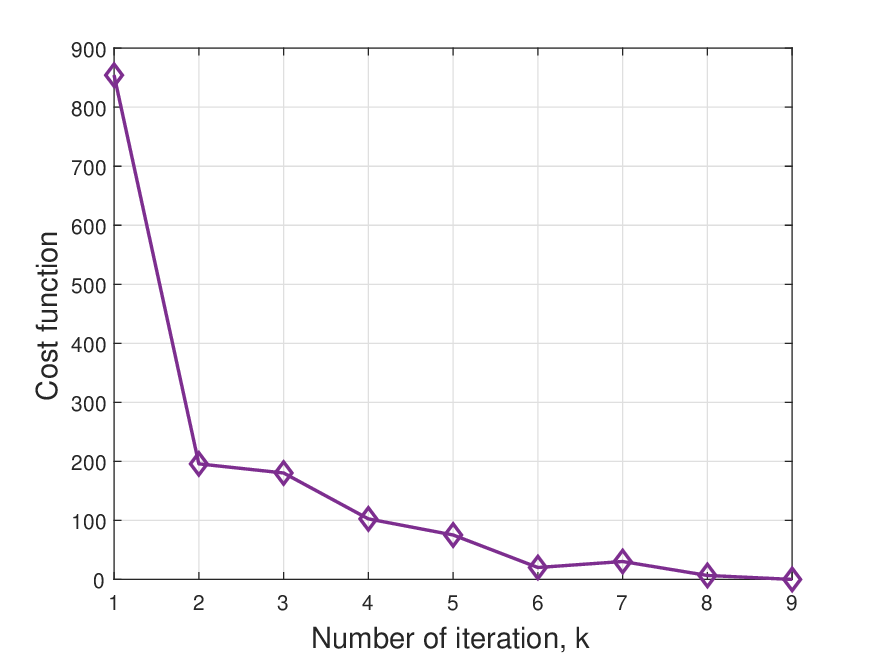}
\vspace{-0.5em}
	\caption{Cost function value versus the number of iterations}
	\label{Convergence}
\end{figure}

\begin{algorithm}[!]
\caption{Proposed Adaptive CMR-ISPS Beamforming}
\begin{algorithmic}[1]
\STATE \textbf{}{Input:}\:Array received data vector $\lbrace \mathbf{{x}}(k) \rbrace_{k=1}^K $;
\STATE \text{Initialize:}\: Compute the sample covariance matrix by \eqref{SCM}; $\mathbf{w}_0=[1, 1, \cdots, 1]^\mathrm{T}$
\STATE Compute the DoA of interference based on \eqref{correlation estimator};
\STATE Calculate $B_{\theta_l}$ and $b_l$ from \eqref{Theta_l} and \eqref{ni};
\STATE Define $Q_{\text{ME}_L}=\bigcup \limits_{l=1}^{L}\{b_l\leq b \leq b_l+{B_{\theta_l}}-1 \}$
\STATE Estimate the desired signal , $\hat{\mathbf{a}}_0$ using \eqref{ahat};
\STATE Set $\mathbf{g}_0= \nabla  J(\mathbf{w}_0)$ and 
$\mathbf{d}_0=-\mathbf{g}_0$;
\STATE Set $t \gets 0$;
\WHILE{$\Vert \nabla  f(\mathbf{w}_t) \Vert > \mathrm{tol.}$}
\STATE Determine the step-size $\mu_t=-{\mathbf{d}_t^\mathrm{T} \mathbf{g}_t } / {\mathbf{d}_t^\mathrm{T}\hat{\mathbf{R}}_{\mathrm{ME}_L} \mathbf{d}_t }$ 
\STATE $\mathbf{w}_{t+1}=\mathbf{w}_t+\mu_t \mathbf{d}_t$;
\STATE Compute $\mathbf{r}_t=\sum_{i=1}^{Q_{\text{ME}_L}}\hat{P}_\mathrm{ME}(\theta_i) \Big(\mathbf{a}^\mathrm{H}(\theta_i)   \mathbf{w}_t\Big)\mathbf{a}(\theta_i)\Delta\theta$
\STATE Define $ \nabla  f(\mathbf{w}_{t+1})=\mathbf{r}_t+\hat{\mathbf{a}}_0 $
\STATE $\mathbf{g}_{t+1}=\nabla  f(\mathbf{w}_{t+1})$
\STATE Determine  $\beta_t={\mathbf{g}_{t+1}^\mathrm{T} (\mathbf{g}_{t+1}-\mathbf{g}_{t})} / {  \parallel \mathbf{g}_{t} \parallel^2}$;
\STATE Set $\mathbf{d}_{t+1}=-\mathbf{g}_{t+1}+\beta_k\mathbf{d}_t$;
\STATE Set $t \gets t+1$;
\ENDWHILE 
\STATE \textbf{Output:}\: Proposed beamformer, $\mathbf{w}_\text{CMR-ISPS}$
\end{algorithmic}
\end{algorithm}
{From a complexity point of view, the weight vector is computed by the proposed iterative CMR-ISPS technique with the cost of $\mathcal{O}(kQ_{\text{ME}_L}M)$, without the need for the inverse of the INC matrix. It should be noted that the detailed complexity comparisons and the value of all parameters are given in \ref{complexity Exp}.}\\
Fig.~\ref{Convergence} shows the convergence of the iterative CMR-ISPS technique in terms of the average of the optimal value found by the algorithm over iterations for SNR=10 dB and T=50. It can be observed that the proposed algorithm converges to the global optimum in about 8 iterations. 
\vspace{-1.1em}
\section{Analysis}\label{sec7}

In this section, we analyze the Capon and the maximum entropy INC matrices and the computational complexity of the proposed CMR-ISPS is detailed.

\subsection{Analysis of the INC Matrix Reconstruction Based on Capon and Maximum Entropy} \label{Analysis}
The accuracy of the power spectrum estimate is clearly the most important issue with INC reconstruction. Inaccuracies in the power spectral estimate result in distorted angular positions and powers of interfering signals, which eventually lead to insufficient interference suppression. An approach in \cite{gu2012robust} introduced the use of the Capon spectrum estimation method to reconstruct the INC matrix. Despite the numerous papers that rely on INC reconstruction based on the Capon estimator \cite{capon1969high}, it was never demonstrated that the reconstruction is, at least roughly, equal to the INC matrix. Although such a result is intuitively expected, a full analytical proof is not tractable, even if the theoretical $\mathbf{R}_{\text{in}}$ replaces the estimated $ \hat{\mathbf{R}}_{\text{in}} $. \\ \indent In the following discussion, we present a justification for the reconstruction of the approaches based on the Capon and the proposed maximum entropy methods, {where the interference power is much larger than the noise power ($\sigma^2_l \gg \sigma^2_\mathrm{n}$)}. {We also stress that, due to its mathematical tractability and simplicity, only the analysis for a single interferer is performed, whereas the proposed algorithm is intended for scenarios with multiple interferers which is shown in the simulation section.}\\
Moreover, it has been proved that the optimal beamformer weight of \eqref{optimal wegight vector} does not change the optimal output SINR when the INC matrix $\mathbf{R}_{\text{(Cap)in}}$ is replaced by the theoretical $\mathbf{R}_{tc}=\sigma^2_\mathrm{n}\mathbf{I}_M+\sigma^2_l\mathbf{a}_l\mathbf{a}_l^\mathrm{H}$ in \cite{van2004detection}. The  INC matrix, $\mathbf{R}_{\text{(Cap)in}}$ in \cite{gu2012robust} is reconstructed as 
\begin{align}\label{Ripn}
\mathbf{R}_{\text{(Cap)in}}=\int_{\Theta_{\text{in}}}\frac{\mathbf{a}(\theta)\mathbf{a}^\mathrm{H}(\theta)}{\mathbf{a}^\mathrm{H}(\theta)\mathbf{R}^{-1}_{tc}\mathbf{a}(\theta)}d\theta \approx \sum_{j=1}^Q \frac{\mathbf{a}(\theta_j)\mathbf{a}^\mathrm{H}(\theta_j)}{\mathbf{a}^\mathrm{H}(\theta_j)\mathbf{R}^{-1}_{tc}\mathbf{a}(\theta_j)} \Delta \theta_j
\end{align}
Based on this approximation and the results in \eqref{denominator} and \eqref{norm}, the summation can be written as
\begin{align} \label{sum Capon}
\mathbf{R}_{\text{(Cap)in}}\simeq&\frac{\sigma^2_\mathrm{n}}{M}\sum_{\theta_j \neq \theta_l}\mathbf{a}(\theta_j)\mathbf{a}^\mathrm{H}(\theta_j)\Delta\theta_j+\big(\frac{\sigma^2_\mathrm{n}+M\sigma^2_l}{M} \big)\mathbf{a}_l\mathbf{a}_l^\mathrm{H}\Delta\theta_j \nonumber\\ =&\frac{\sigma^2_\mathrm{n}}{M}\sum_{\theta_j\in\Theta_{\text{in}}}\mathbf{a}(\theta_j)\mathbf{a}^\mathrm{H}(\theta_j) \Delta \theta_j +\sigma^2_l\mathbf{a}_l\mathbf{a}_l^\mathrm{H}\Delta\theta_j.
\end{align}
Since $\Theta_{\text{in}} \cup \Theta_s=[-\pi,\pi]$ and the size of the set $\Theta_s$ is much smaller than the size of $\Theta_{\text{in}}$ (measuring "size" in terms of the sum of lengths of the intervals that compose the sets, i.e., the Borel measure), it can be shown that 
\begin{align}\label{App integral}
\int_{\Theta_{\text{in}}} \mathbf{a}(\theta)\mathbf{a}^\mathrm{H}(\theta) d \theta \approxeq \int_{[-\pi,\pi]}\mathbf{a}(\theta)\mathbf{a}^\mathrm{H}(\theta) d \theta=2\pi \mathbf{I}_M,
\end{align}
so that the summation can also be approximated by (\ref{App integral}).  The same considerations about the size of $\Theta_s$ and $\Theta_{\text{in}}$ also allow us to approximate, $\Delta\theta_j \approx \frac{2\pi}{M}$, resulting in
\begin{align}\label{R8}
\mathbf{R}_{\text{(Cap)in}}\simeq \frac{2 \pi}{M} \big[ \sigma^2_\mathrm{n}\mathbf{I}_M+\sigma^2_l\mathbf{a}_l\mathbf{a}_l^\mathrm{H} \big].
\end{align}
A comparison of (\ref{R8}) with the original INC matrix in (\ref{R capon}) shows that the reconstruction only multiplies the true matrix by a factor $\frac{2\pi}{M}$.
On the other hand, by using the same procedure in \eqref{sum Capon} the INC matrix of maximum entropy can be given as 
\begin{align} \label{sum Entropy}
\mathbf{R}_{\text{(ME)in}}\simeq&\sigma^2_\mathrm{n}\sum_{\theta_j \neq \theta_l}\mathbf{a}(\theta_j)\mathbf{a}^\mathrm{H}(\theta_j)\Delta\theta_j+\frac{\big(\sigma^2_\mathrm{n}+M\sigma^2_l \big)^2}{\sigma^2_\mathrm{n}}\mathbf{a}_l\mathbf{a}_l^\mathrm{H}\Delta\theta_j \nonumber\\ 
\simeq & \frac{2\pi}{M} \big[\sigma^2_\mathrm{n}\mathbf{I}_M+ \dfrac{(M^2\sigma^4_l)}{\sigma^2_\mathrm{n}}\mathbf{a}_l\mathbf{a}_l^\mathrm{H}     \big].
\end{align}
The above equation implies that the reconstructed INC by maximum entropy preserves the interference components while the power of each interference is reinforced. Compared to the Capon-based INC in \eqref{R8}, it further enhances the quality of the INC matrix.

\subsection{Computational Complexity} \label{complexity Exp}

Given an array of $M$ elements, $T$ (number of snapshots), and $Q_{\text{ME}_L}$ (union of the small angular sectors of the interferences), the main difference here from previous works lies in the fact that the integral (\ref{C}) is approximated by a summation (\ref{proposed Rin summation}), which requires a complexity of $ \mathcal{O}(Q_{\text{ME}_L}M^2)$ to synthesize the narrowband signal power accurately. However, in the computation of \eqref{Grad with w} and for CMR-ISPS  in (\ref{final W}), we avoid computing expensive $O(M^2)$ outer products, so CMR-ISPS requires $\mathcal{O}(Q_{\text{ME}_L}M)$ for steps (7,9,10,12,13 and 15) while steps in (11 and 14) need $\mathcal{O}(M)$ complexity.  Since CMR-ISPS iterates $k$ times to find the best step size, $\mu_k$. Hence, the computational cost of CMR-ISPS is only $\mathcal{O}(kQ_{\text{ME}_L}M)$, while computing the beamformer without the need for the inverse of the INC matrix and $k \le M \le Q_\text{ME} \ll Q$. In the proposed method $Q_\text{ME}=20$ and $k=8$ is considered\\
\indent To reconstruct the INC matrix, the CMR-SV and CMR-EST beamformers have a complexity of $\text{max}( \mathcal{O}(QM^2), \mathcal{O}(M^{3.5}))$, while the CMR-CC has the complexity $ \mathcal{O}(QM^2$). The CMR-SUB beamformer has a computational complexity of $ L \times \text{max} \big( \mathcal{O}(Q_lM^2),\mathcal{O}(QM^2 ) \big)$ where $Q_l$ denotes the number of samples in the small angular sector of the interference region and $Q$ is the number of uniform samples in the range of the interference-plus-noise region. The CMR-SPSS beamformer has the complexity of $\mathcal{O}(M^3)$ and the CMR-OS has the computational complexity of $\text{max}(\mathcal{O}(JM^2), \mathcal{O}(QM^2))$ where $J$ is the number of sampling point for noise region and finally, the CMR-SVEST beamformer has the complexity of $\text{max}( \mathcal{O}(Q_{i}M^2), \mathcal{O}(M^{3.5}))$ where $Q_i$ depicts the number of uniform samples in the interference-plus-noise region given by the authors.  In all simulation results, the parameters $Q=200$, $Q_l=40$, and $Q_i=1002$ are considered. Note that, for each algorithm in the simulations we employed the given values for parameters in the papers or lowest quantities that result in the best performance because some references did not provide values for the parameter $Q$. \\

\begin{figure}[h]
	\centering
	\includegraphics[width=2.8in,height=2.1in]{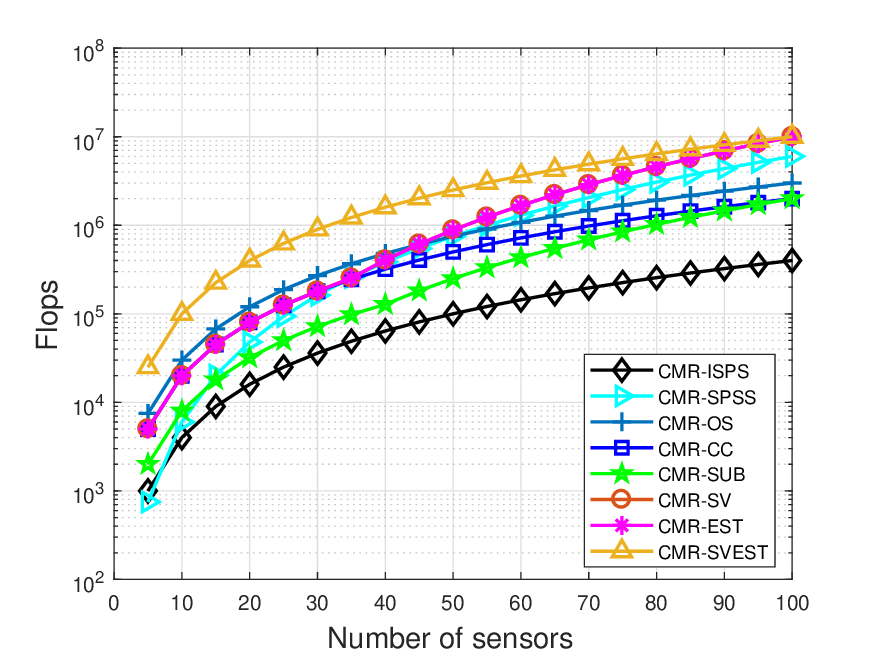}
\vspace{-0.5em}
	\caption{Complexity comparison}
	\label{Complexity}
\end{figure}

\indent As an example to show the computational complexity of the proposed method and other compared methods, we ran all algorithms using MATLAB 2017a on a Windows 10 laptop with dual-core 1.9 GHz Intel Core i3 CPU and 3.36 GB memory. the results are demonstrated in Table~\ref{tab:expcond}. To do this properly, we closed all background processes in our computer (internet, e-mail, music, all unrelated software), and restricted MATLAB to use a single processor at all times. Furthermore, it should be noted that these measurements have been done based on the parameters given in the simulation section and the scenario that there is no mismatch between the actual and assumed steering vector. The exact signal steering vector is known.

\begin{table}[h] \label{com}
  \centering
  \caption{Computational time}
  \label{tab:expcond}
  \begin{tabular}{c |c }
    \hline \hline
    Beamformers  & Execution time (s)    \\ 
    \hline \hline
CMR-ISPS   &$ 0.0025 $   \\
CMR-SPSS   &$ 0.0065 $   \\ 
CMR-SUB   &$ 0.0322 $   \\
CMR-CC   & $0.0403$     \\
CMR-OS  &$ 0.0747 $   \\
CMR-EST   &$ 0.7186 $    \\
CMR-SV  &$ 0.7207 $   \\
CMR-SVEST  &$ 0.8669$    \\
    \hline
\end{tabular}
\end{table}

The computational complexity analysis is demonstrated in Fig.~\ref{Complexity}. We measure the total number of additions and multiplications (i.e., flops) in terms of the number of sensors $M$ performed for each snapshot for the proposed algorithm and the other tested ones. Note that the CMR-SVEST method has a highly-variant computational complexity in different snapshots, due to the online optimization program based on the  estimation. However, it needs more uniform sampling points for interference plus noise angular sector which is given $Q_i=1002$. Therefore, it needs high computational complexity. The complexity of the CMR-SV and CMR-EST suffers from solving the quadratically constrained quadratic program to reconstruct the INC matrix. On the other hand, although the CMR-SPSS requires less computation for 30 number of sensors compared to the other methods, the complexity of this method is increased by more sensors. Moreover, the CMR-OS method needs more sampling points for the noise angular sector ($J$) in order to reconstruct the INC matrix As can be seen, the proposed CMR-ISPS algorithm has lower complexity than the other algorithms.

\section{Simulations}\label{sec8}
Consider a uniform linear array with $M = 10$ omnidirectional antenna elements spaced one-half wavelength apart. The additive noise is modeled as a complex Gaussian zero-mean spatially and temporally white process with unit variance in each sensor. Extensions to large arrays \cite{mmimo,wence} and communications problems \cite{spa,mfsic,bfidd,listmtc,dynmtc,dynovs,jointmtc} are straightforward. The presumed direction towards the SOI is assumed to be $\bar{\theta}_0=10^\circ$, while two interfering sources are assumed to impinge on the sensor array from the directions $\bar{\theta}_1=20^\circ$ and $\bar{\theta}_2=-40^\circ$, respectively. For each interfering signal, a single sensor's interference-to-noise ratio (INR) equals 30dB. Besides, $ {\Theta}_s=[{\bar{\theta}_0}-4^\circ,{\bar{\theta}_0}+4^\circ]$  is the angular sector of the SOI and the region of interference signals is $\Theta_l=\Theta_1 \cup \Theta_2=[\bar{\theta}_1-4,\bar{\theta}_1+4]\cup [\bar{\theta}_2-4, \bar{\theta}_2+4]$. In each scenario, 100 Monte-Carlo runs have been performed to produce the results. The proposed CMR-ISPS method is compared with the beamformer in \cite{yuan2017robust} (CMR-SUB),  the reconstruction-estimation based beamformer in \cite{gu2012robust} (CMR-EST), the beamformer in \cite{zhu2020robust} (CMR-OS), the beamformer in \cite{zhang2016interference} (CMR-SPSS), the beamformer in \cite{zheng2018covariance} (CMR-SV), the beamformer in \cite{chen2015robust} (CMR-CC) and the beamformer method in \cite{khabbazibasmenj2012robust} (CMR-SVEST). In \cite{zhang2016interference}, the reference angle is $\alpha_0=0^\circ$ and the null broadening parameter is $\Delta=\sin^{-1}(2/M)$. In the CMR-SV beamformer, the upper bound of the norm of the  mismatch is set to $ \epsilon=\sqrt{0.1} $. The number of  dominant eigenvectors is chosen as 7 in CMR-OS. The energy percentage $\rho$ is set to 0.9 in CMR-SUB. The CVX toolbox \cite{grant2008cvx} is used to solve convex optimization problems. It should be noted that, in all scenarios, the input SNR and the optimal output SINR  are defined as $\mathrm{SNR} = {\sigma^2_0}/{\sigma^2_n}$ and $\mathrm{SINR}_\mathrm{opt}=\sigma^2_0 | \mathbf{a}_0^\mathrm{H} \mathbf{R}_{\text{in}}^{-1}{\mathbf{a}}_0 |$, respectively. It should be noted that the utilized DoAs for compared algorithms are the same as the scenarios which are given in the corresponding papers.
\begin{figure}[!]
\centering
\subfloat[Output SINR versus SNR]{\label{SNR_Look} \includegraphics[width=2.8in,height=2.1in]{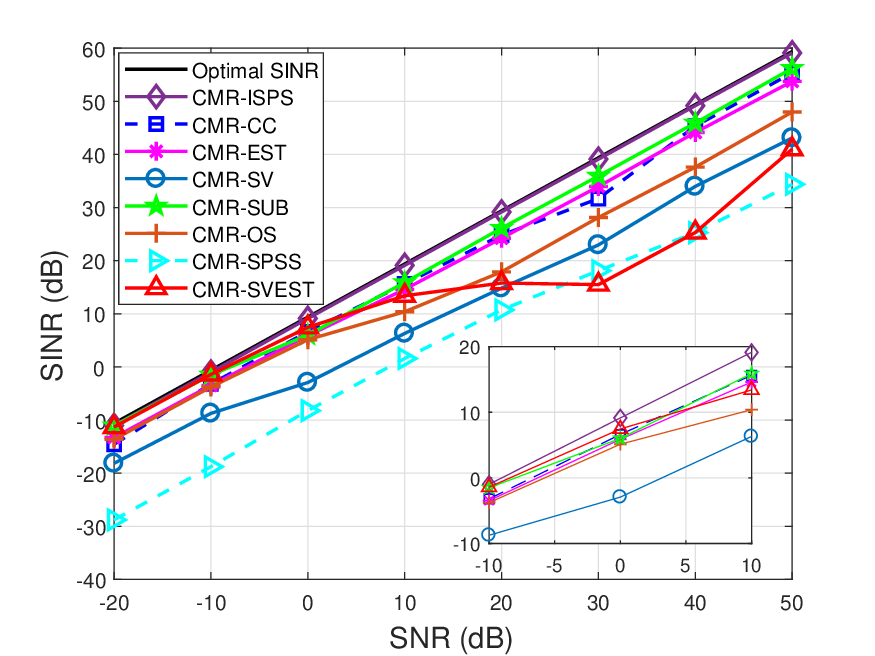}}
\qquad
\subfloat[Output SINR versus number of snapshots]{\label{Snap_Look} \includegraphics[width=2.8in,height=2.1in]{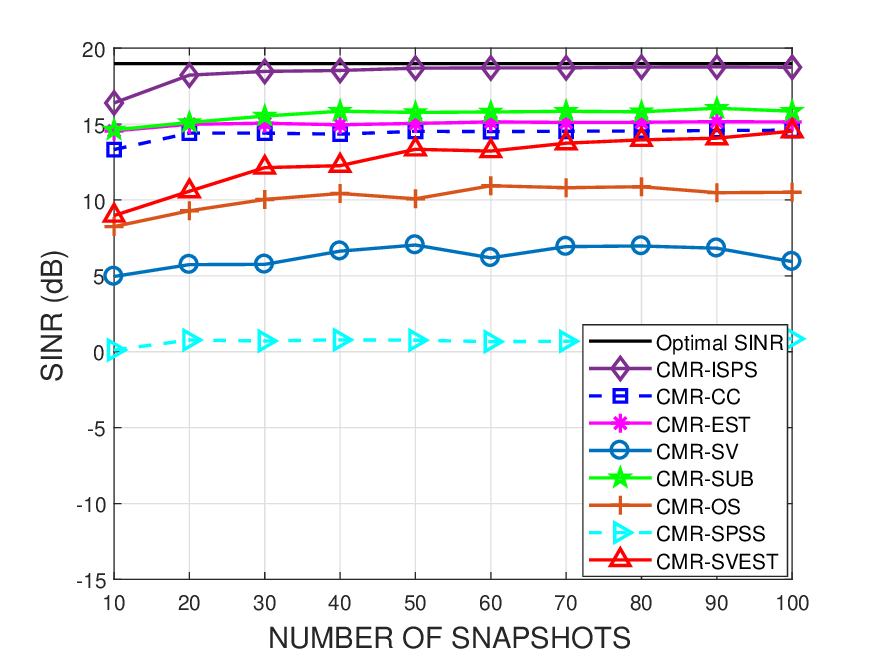}}
\caption{Look Direction Mismatch}
\label{fig}
\end{figure}

\subsection{Mismatch due to look direction errors}
In this example, the performance of all the above methods is evaluated in terms of mismatch due to looking direction error. Assume that the error of all the signals is subject to a uniform distribution in $[-4^\circ; 4^\circ] $ {space} for each simulation. In other words, the angles of the SOI and interference signals are uniformly distributed in $[6^\circ;14^\circ]$, $[16^\circ;24^\circ]$ and $[-44^\circ; -36^\circ]$ respectively. Fig.~\subref*{SNR_Look} demonstrates the output SINR of the tested methods versus the input SNR with $T = 50$ snapshots. The proposed CMR-ISPS method has excellent performance compared with other beamformers. Also, it should be noted that since the procedure of the DoA estimation enables the proposed CMR-ISPS method to adapt itself to the motion of interference by determining the angular region, CMR-ISPS is more effective against look direction errors. 
It is evident that the performance of the CMR-OS and CMR-SV is sensitive when the DoA of the desired signal and interference are close and their performance suffers from random look direction mismatch. On the other hand, the CMR-EST and CMR-SUB show acceptable performance {while the CMR-SVEST has good performance up to 0 dB and its result is degraded for high SNRs}. Fig.~\subref*{Snap_Look} displays the output SINR of the same methods versus the number of snapshots for an SNR of 10 dB. It is evident that the proposed CMR-ISPS algorithm realizes the most outstanding performance under different numbers of snapshots, compared with the other three CMR-SUB, CMR-EST, and CMR-CC beamformers. {On the other hand, the CMR-SVEST beamformer obtains a better performance with snapshots increment.} However, the performance of CMR-OS, CMR-SV, and CMR-SPSS is not improved with training data snapshots increment. The existing reconstruction-based CMR-OS, CMR-SV, and CMR-SPSS methods achieve better performance improvement under the assumption that the interferences are not close to the signal angular sector. However, their performance is degraded under the close DoA scenario.
\begin{figure}[!]
\centering
\subfloat[Beampattern for Close DoAs]{\label{Beam close} \includegraphics[width=2.8in,height=2.1in]{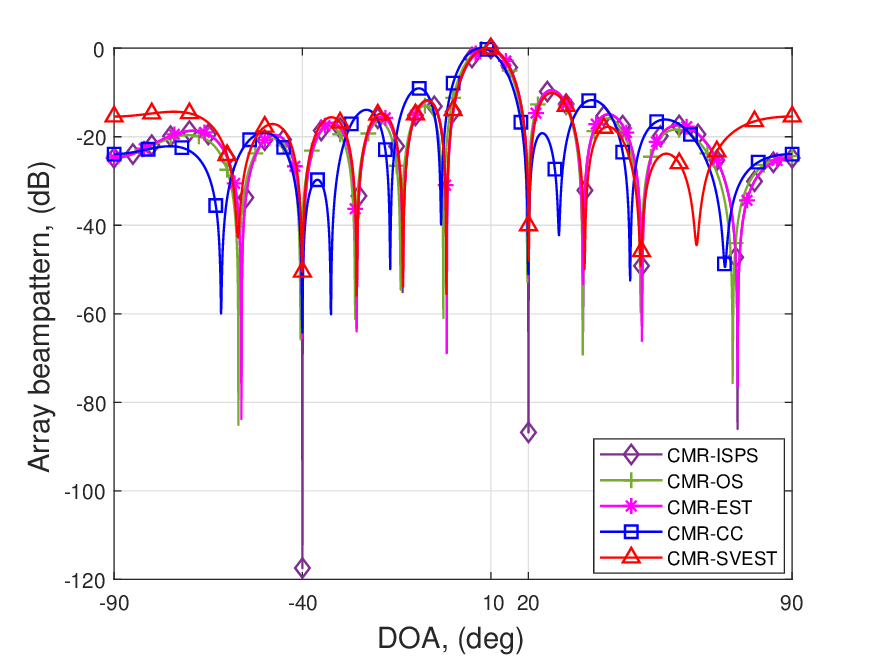}}
\qquad
\subfloat[Beampattern for far DoAs]{\label{Beam far} \includegraphics[width=2.8in,height=2.1in]{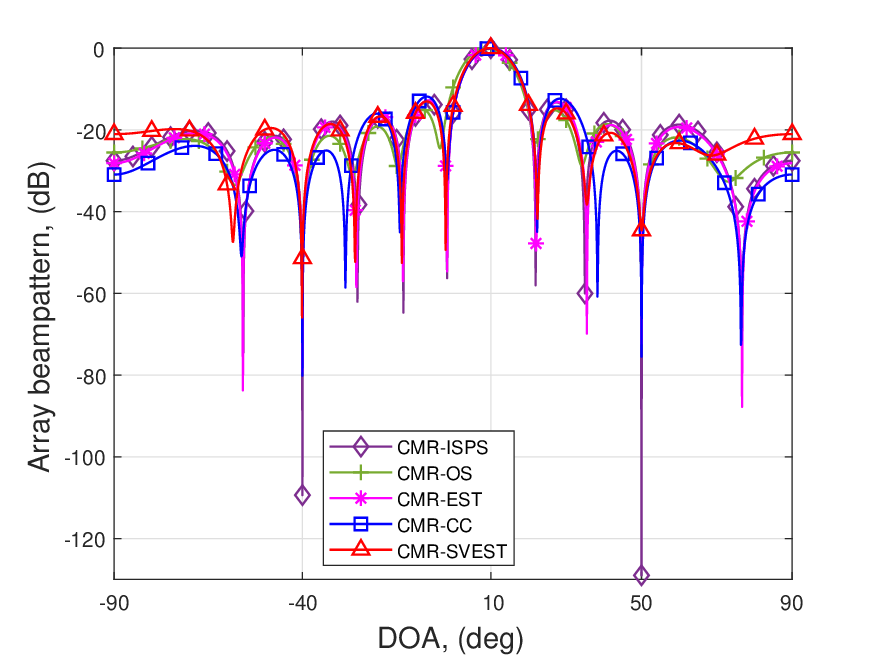}}
\caption{The beampattern of the beamformers}
\label{fig}
\end{figure}

{\subsection{Array beampattern}

In the second example, the beampattern of the proposed CMR-ISPS beamformer is compared with the CMR-OS, CMR-EST, CMR-CC, and CMR-SVEST beamformers, when the number of snapshots is fixed at $T = 100$. In this example, the presumed direction towards the SOI is assumed to be $\bar{\theta}_0=10^\circ$ with SNR of 10 dB which has no array pointing error. The interferences are considered in two scenarios. First, the interference direction $[20^\circ, -40^\circ]$ is close to the desired signal, and then the interference direction $[-40^\circ, 50^\circ]$ is far from the desired signal. Considering the influence of close DoAs in Fig.~\subref*{Beam close}, it can be seen that, although all tested methods are able to point the main lobe to the actual direction of the desired signal DoA, the proposed CMR-ISPS beamformer can suppress interferers that are close to the direction of the SOI by producing notches in the directional response of the array with deep depths. Fig.~\subref*{Beam far} depicts the performance of the proposed and examined methods in the scenario of sufficiently far DoAs for desired signal and interferences. We find that all methods have nearly the same main lobe beam width. Also, we can see that in the proposed method the Interference signals are efficiently nulled compared to other tested methods.}

\subsection{Mismatch due to array geometry errors}

\begin{figure}[h]
\centering
\graphicspath{{},{}}
\subfloat[Output SINR versus SNR]{\includegraphics[width=2.8in,height=2.1in]{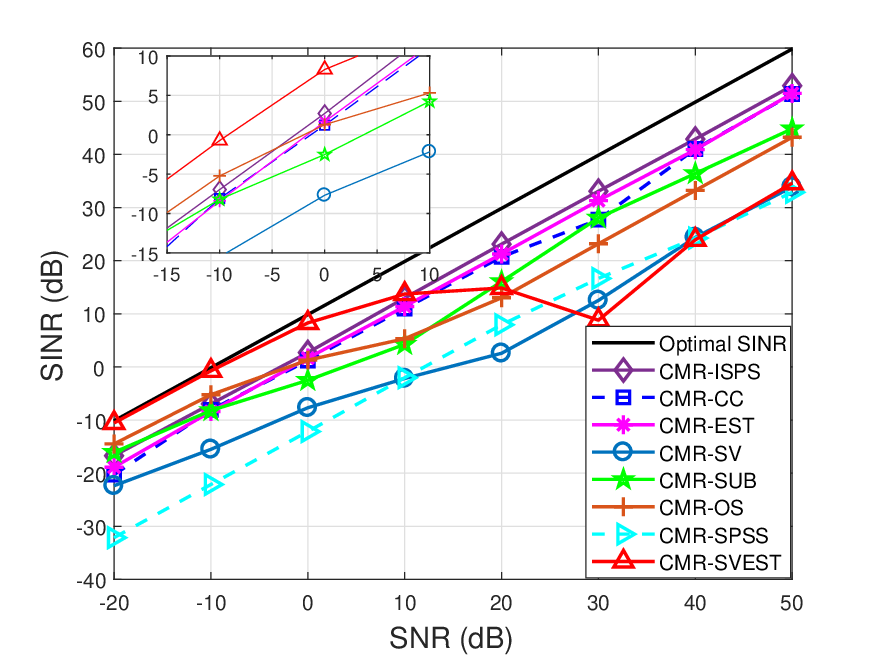}
\label{SNR_Geo}}
\qquad
\subfloat[Output SINR versus number of snapshots ]{\includegraphics[width=2.8in,height=2.1in]{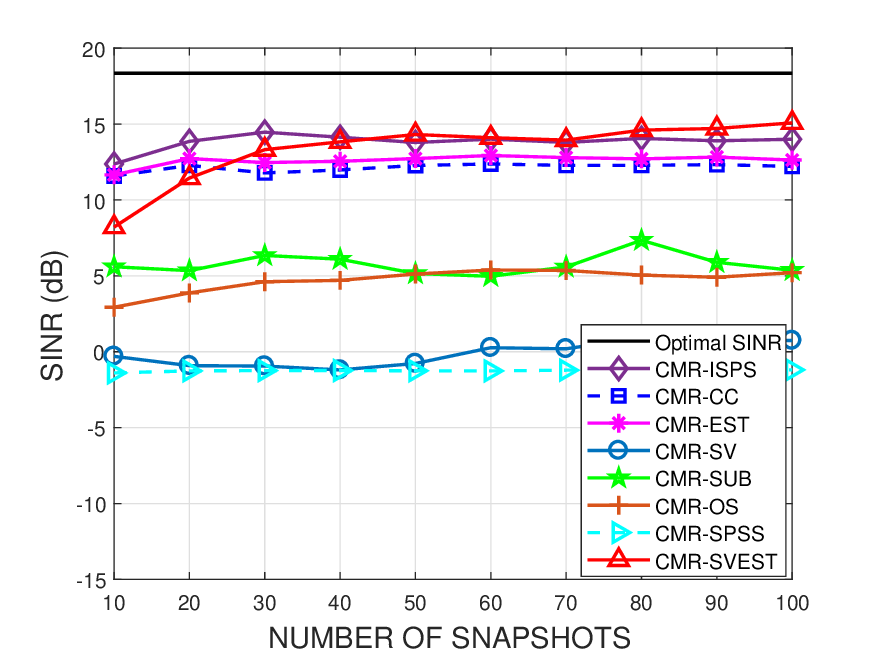}
\label{Snap_Geo}}
\caption{Array Geometry Error}
\label{Array Geometry Error}
\end{figure}

In the third example, we investigate how the presence of array geometry errors affects the beamformer output SINR. Let the sensor array perturbations be caused by errors in the sensor element positions which are drawn uniformly from the interval $[-0.05,0.05]$ measured in wavelengths. Fig.~\subref*{SNR_Geo} displays the output SINR of beamformers versus the input SNR under the condition of the number of snapshots $T = 50$. As shown in Fig.~\subref*{SNR_Geo}, it can be found that {although CMR-SVEST achieves the best performance for SNRs from -20 dB to 10 dB,} the proposed CMR-ISPS method performs best among the tested beamformers. Meanwhile, the proposed beamformer obtains almost optimal output SINR as input SNR increases, which also highlights the proposed beamformer has better effectiveness against array geometry error. Although at low SNRs, CMR-OS shows better performance, CMR-ISPS and CMR-SUB have similar performance, and both of them outperform other beamformers. However, CMR-SUB degrades its performance in SNR greater than -10 dB. Fig.~\subref*{Snap_Geo}, {depicts the output SINR of tested beamformers versus the number of snapshots for the fixed SNR = 10 dB.} We notice that CMR-ISPS performs better than the other beamformers which benefit from the more accurate  estimation and INC matrix reconstruction while the CMR-SVEST beamformer performance is improved with the number of snapshots. However, the CMR-SUB beamformer is sensitive to sensor displacement errors.

\vspace{-0.5em}
\subsection{Mismatch due to amplitude and phase perturbations errors}

\begin{figure}[h]
\centering
\graphicspath{{},{}}
\subfloat[Output SINR versus SNR]{\includegraphics[width=2.8in,height=2.1in]{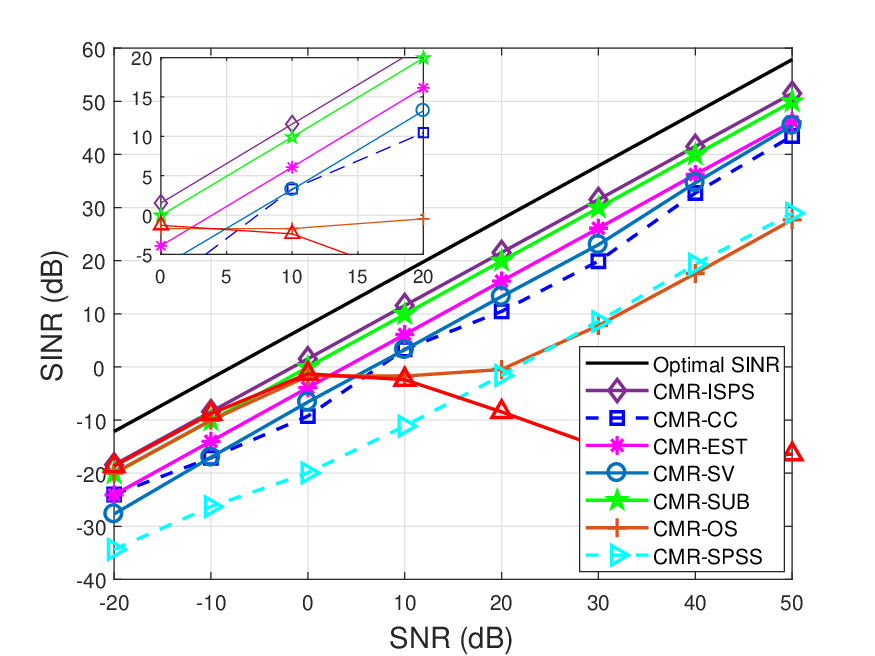}
\label{SNR_Gain}}
\qquad
\subfloat[Output SINR versus number of snapshots ]{\includegraphics[width=2.8in,height=2.1in]{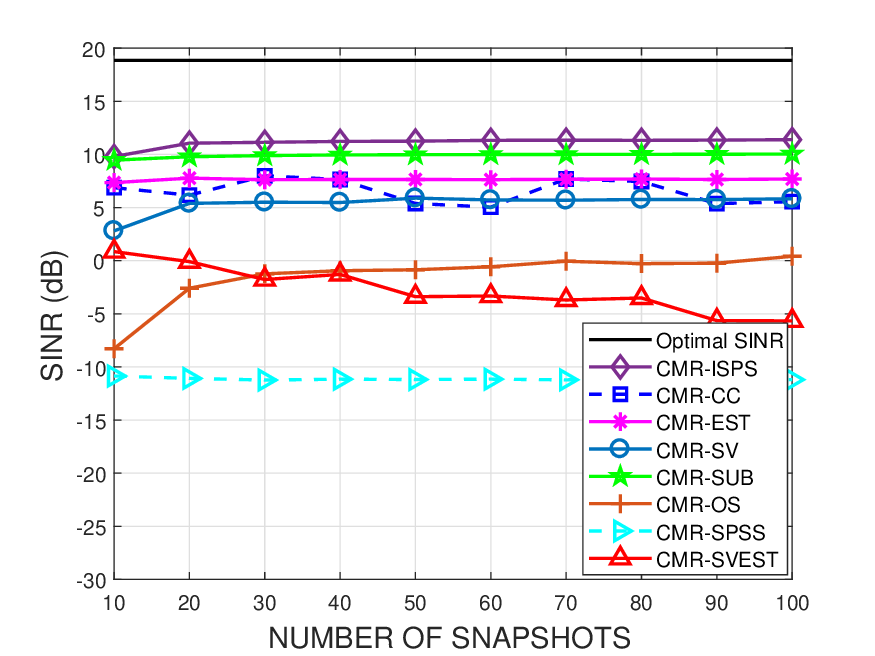}
\label{Snap_Gain}}
\caption{Gain and Phase Error}
\label{Gain and Phase Error}
\end{figure}

Here, the effects of arbitrary amplitude and phase perturbation errors of the array channel are analyzed. We assume that the calibration error is caused by gain and phase perturbations in each sensor which is drawn from the corresponding random generators $\mathcal{N}(1,0.05^2)$ and $\mathcal{N}(0,(0.025\pi)^2)$. The output SINR of the methods under comparison versus the input SNR for the number of snapshots $T=50$ is shown in Fig.~\subref*{SNR_Gain}. It can be seen that CMR-ISPS outperforms the other beamformers for the whole range of SNRs. {Moreover, the performance of the CMR-SVEST beamformer is degraded severely in the case of gain and phase mismatches while the CMR-OS degrades its performance for SNRs greater than 10 dB.} Fig.~\subref*{Snap_Gain} corresponds to the performance curves versus the number of snapshots at fixed SNR=10 dB. We note that CMR-ISPS enjoys better performance than the others. In this uncalibrated array scenario, since prior knowledge about the steering vector is imprecise, the CMR-EST reconstruction method degrades its performance compared to other approaches. On the other hand, CMR-SV use effectively the information of the uncertainty set about the steering vector {and the performance of CMR-SVEST is degraded for snapshots}. 

\vspace{-0.5em}
\subsection{Mismatch Due to Incoherent Local Scattering}

\begin{figure}[h]
\centering
\graphicspath{{},{}}
\subfloat[\small Output SINR versus SNR ]{\includegraphics[width=2.8in,height=2.1in]{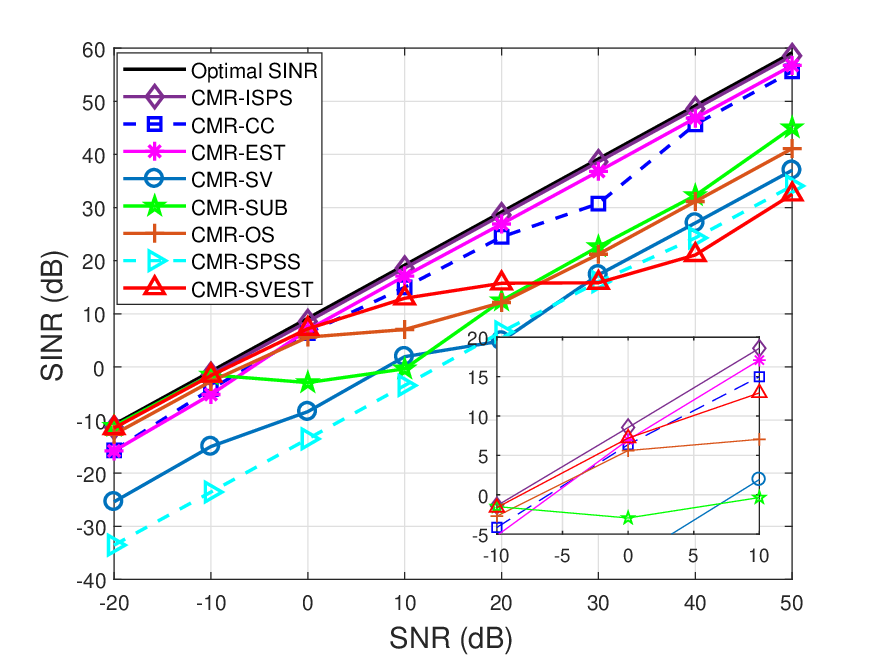}
\label{SNR_Inco}}
\hfil \\ \vspace{-1em}
\subfloat[\small Output SINR versus number of snapshots ]{\includegraphics[width=2.8in,height=2.1in]{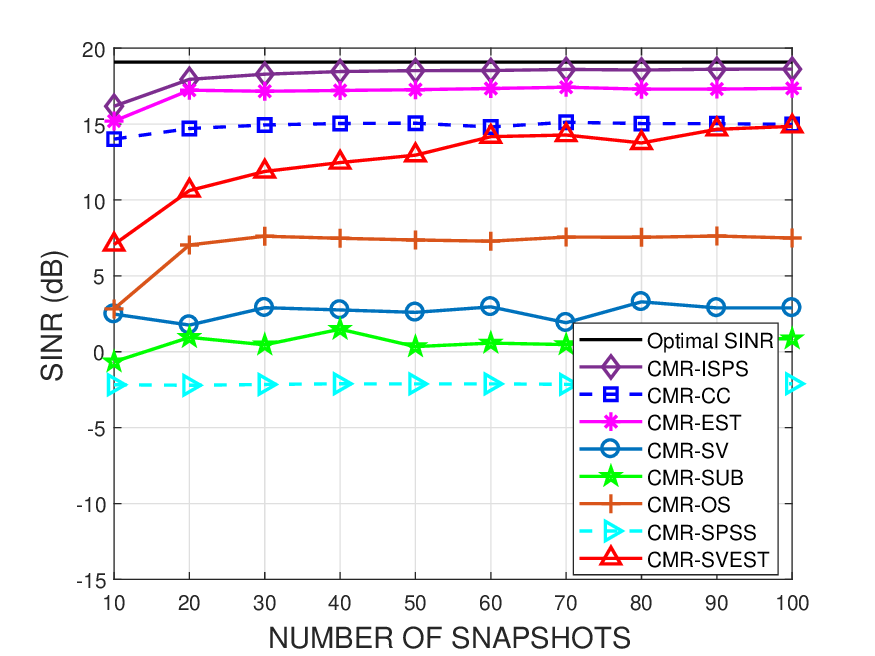}
\label{Snap_Inco}}
\caption{\small Incoherent Local Scattering Mismatch}
\label{Incoherent local scattering mismatch}
\end{figure}

In this example, a scenario with an incoherent local scattering of the SOI is considered. The signal is assumed to have a time-varying spatial signature, and the steering vector of the SOI is modeled as $\mathbf{a}(t)=s_0(t)\mathbf{{a}}(\theta_0)+\sum_{p=1}^4 s_p(t)\mathbf{{a}}(\theta_p)$,
where $s_0(t)$ and $s_p(t)$ $(p=1,2,3,4)$ are independently and identically distributed (i.i.d.) zero-mean complex Gaussian random variables are independently drawn from a random generator. The DoAs $\theta_p$ $(p=0,1,2,3,4)$ are independently normally distributed in $ \mathcal{N} (\bar{\theta}_0,4^\circ)$ in each simulation run. Note that $\theta_p$ changes from run to run, but remains fixed from snapshot to snapshot. At the same time, the random variables  $s_0(t)$ and $s_p(t)$ change not only from run to run but also from snapshot to snapshot. This corresponds to the case of incoherent local scattering \cite{besson2000decoupled}, where the SOI covariance matrix $\mathbf{R}_\text{s}$ is no longer a rank-one matrix. In this situation, the optimal output SINR is given by a more general form $\mathrm{SINR}_\mathrm{opt}={\mathbf{w}^H \mathbf{R}_\mathrm{s}\mathbf{{w}} } / {\mathbf{w}^H \mathbf{R}_\mathrm{in}\mathbf{w}}$,
which is maximized by the weighted vector $\mathbf{{w}}_\mathrm{opt}= \mathcal{P} \lbrace \mathbf{{R}}_\mathrm{in}^{-1} \mathbf{R}_\mathrm{s} \rbrace$, where $\mathcal{P} \lbrace \cdot \rbrace$ represents the principal eigenvector of a matrix. \\ Fig.~\subref*{SNR_Inco} depicts the output SINR of the tested beamformers versus the input SNR while the number of snapshots is fixed at $T=50$. It is observed that CMR-ISPS has the highest output SINR, which is almost with the optimal beamformer Meanwhile, the performance of CMR-SUB and CMR-OS degrades with the increase of SNR because the real steering vector is magnified due to the introduced incoherent local scattering error. Considering the influence of the change in the snapshots with fixed 10 dB SNR, the performance of the tested beamformers is described in Fig.~\subref*{Snap_Inco}, where all the beamformers perform stably as the snapshots change. At the same time, there is an obvious difference between CMR-ISPS and other tested methods, where CMR-ISPS almost reaches the optimal beamformer as the number of snapshots increases.

\vspace{-.3em}
\section{Conclusion}\label{sec9}

In this paper, we have introduced a cost-effective approach to robust adaptive beamforming, denoted by CMR-ISPS, based on the INC matrix reconstruction and desired signal  estimation. The central idea is that each interference DoA is estimated during the period in which snapshots are taken while its power is obtained by using the maximum entropy spectral estimator. 
Moreover, we estimated the steering vector of SOI by utilizing the SOI covariance matrix which is achieved from the spatial power spectrum. 
Numerical results have demonstrated that CMR-ISPS can achieve better performance against unknown arbitrary-type mismatches. \vspace{-0.6em}

\bibliographystyle{IEEEtran}
\bibliography{Biblio.bib}

\end{document}